%% file: main.tex
\documentclass{article} % For LaTeX2e
\usepackage{iclr2025_conference,times}

\input{math_commands.tex}

\usepackage{hyperref}
\usepackage{url}
\usepackage{graphicx,verbatimbox}  % Scalebox
\usepackage{booktabs}
\usepackage{amssymb}
\usepackage{xcolor}
\usepackage{diagbox}
\usepackage{caption}
\usepackage{url}            % simple URL typesetting
\usepackage{booktabs}       % professional-quality tables
\usepackage{amsfonts}       % blackboard math symbols
\usepackage{nicefrac}       % compact symbols for 1/2, etc.
\usepackage{microtype}      % microtypography
\usepackage{xcolor}         % colors
\usepackage{graphicx}
\usepackage{amsmath}
\usepackage{amssymb}
\usepackage{booktabs}
\usepackage{makecell}
\usepackage{multirow}
\usepackage{float}
\usepackage{mathtools}
\usepackage{amsmath}

\usepackage[utf8]{inputenc} % allow utf-8 input
\usepackage[T1]{fontenc}    % use 8-bit T1 fonts
\usepackage{algorithm}% http://ctan.org/pkg/algorithms
\usepackage{algorithmicx}% http://ctan.org/pkg/algorithms
\usepackage[noend]{algpseudocode}
\algrenewcommand\alglinenumber[1]{\tiny #1:}
\usepackage{fp}
\usepackage{wrapfig,lipsum,epstopdf}
\usepackage{lipsum}
\usepackage{rotating}
\usepackage{multirow}
\usepackage{pifont}

\usepackage{array, makecell} 
\usepackage{xcolor,colortbl}
\usepackage{comment}
\usepackage{kantlipsum}
\usepackage{tabularx}
\usepackage{amsmath}
\usepackage{diagbox}
\usepackage{fancyhdr}
\usepackage{float}
\usepackage{caption}
\usepackage{graphicx}
\usepackage{subcaption}
\usepackage{etoolbox}
\usepackage{enumitem}
\usepackage{adjustbox}
\usepackage{arydshln} % Required for dashed lines in tables
\setlist{topsep=0pt,noitemsep,topsep=0pt,parsep=0pt,partopsep=0pt}
\definecolor{bgreen}{rgb}{0.0, 0.5, 0.0}
\definecolor{deepskyblue}{rgb}{0.0, 0.75, 1.0}

\newcommand{\papername}{Interactive Structured Image Generation with Cascaded Schrödinger Bridge}
\newcommand{\papernameAbbrev}{ToddlerDiffusion}

\title{{\papernameAbbrev}: {\papername}}

\author{
\quad \quad \quad \quad \quad \quad \quad Eslam Abdelrahman $^1$,
Liangbing Zhao $^1$,
Vincent Tao Hu $^2$, \\
\quad \quad \quad \quad \quad \quad \quad \quad \textbf{Matthieu Cord} $^3$,
\textbf{Patrick Perez} $^4$,
\textbf{Mohamed Elhoseiny} $^1$ \\
\hspace{0.1cm}
\quad \quad \quad \quad \quad \quad \quad \quad \quad \quad 
$^1$KAUST \quad $^2$LMU \quad $^3$ Valeo AI \quad $^4$ Kyutai
}

\iclrfinalcopy % Uncomment for camera-ready version, but NOT for submission.
\begin{document}

\makeatletter
    \let\@oldmaketitle\@maketitle
    \renewcommand{\@maketitle}{\@oldmaketitle
        \myfigure\bigskip}
    \makeatother
    \newcommand\myfigure{%
    % \vspace{-4mm}
    \centering
      \includegraphics[width=\linewidth]{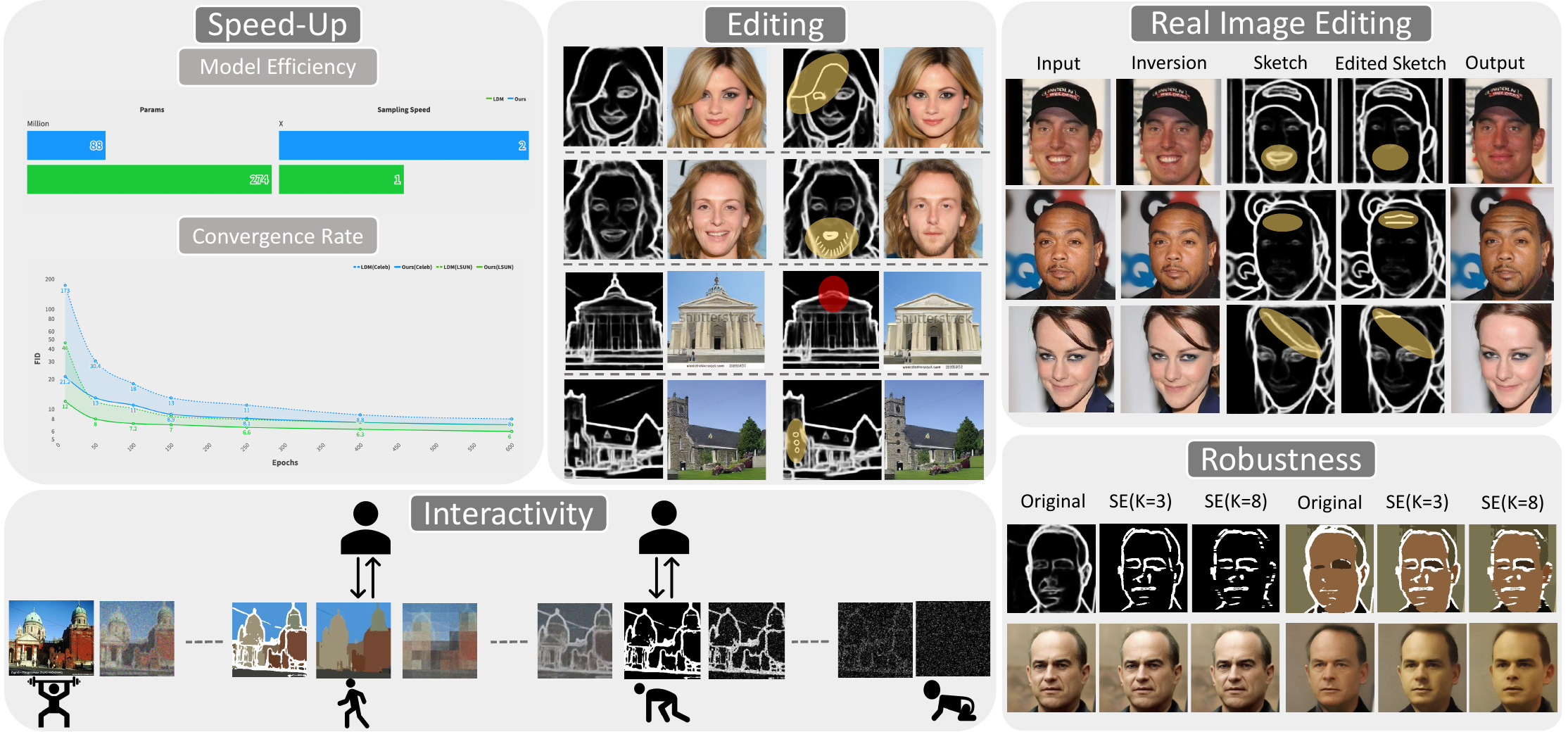}
         % \\
          \centering
          % \vspace{0.5cm}
          \refstepcounter{figure}\normalfont\footnotesize{{Figure~\thefigure: An overview of {\papernameAbbrev} capabilities: 1) Speed-Up: Performing on the bar compared to LDM while being 2$\times$ faster, using 3$\times$ smaller architecture. 2) Interactivity: Producing intermediate outputs, e.g., sketch and palette. 3) Editing: Offering free, robust editing capabilities even for real images. 4) Robustness: Our method is robust against input condition perturbations, e.g., sketch sparsity.} 
          } 
        \label{fig_teaser_fic}
        \vspace{-4mm}
}

\maketitle

\input{sections/0_abstract}
\input{sections/1_Intro}
% \input{sections/3_revisit}
\input{sections/4_methodology}
\input{sections/5_exp_new}
\input{sections/7_conclusion}

\clearpage

\bibliography{main}
\bibliographystyle{iclr2025_conference}

\clearpage
% \appendix
% \section{Appendix}
\input{sections/appendix}

\end{document}

%% file: math_commands.tex
\usepackage{amsmath,amsfonts,bm}

\def\eqref#1{equation~\ref{#1}}
\def\1{\bm{1}}

\DeclareMathAlphabet{\mathsfit}{\encodingdefault}{\sfdefault}{m}{sl}
\SetMathAlphabet{\mathsfit}{bold}{\encodingdefault}{\sfdefault}{bx}{n}

%% file: sections/0_abstract.tex
\begin{abstract}
Diffusion models break down the challenging task of generating data from high-dimensional distributions into a series of easier denoising steps. Inspired by this paradigm, we propose a novel approach that extends the diffusion framework into modality space, decomposing the complex task of RGB image generation into simpler, interpretable stages. Our method, termed {\papernameAbbrev}, cascades modality-specific models, each responsible for generating an intermediate representation, such as contours, palettes, and detailed textures, ultimately culminating in a high-quality RGB image.
Instead of relying on the naive LDM concatenation conditioning mechanism to connect the different stages together, we employ Schr\"odinger Bridge to determine the optimal transport between different modalities.
Although employing a cascaded pipeline introduces more stages, which could lead to a more complex architecture, each stage is meticulously formulated for efficiency and accuracy, surpassing Stable-Diffusion (LDM) performance.
Modality composition not only enhances overall performance but enables emerging proprieties such as consistent editing, interaction capabilities, high-level interpretability, and faster convergence and sampling rate. 
Extensive experiments on diverse datasets, including LSUN-Churches, ImageNet, CelebHQ, and LAION-Art, demonstrate the efficacy of our approach, consistently outperforming state-of-the-art methods.
For instance, {\papernameAbbrev} achieves notable efficiency, matching LDM performance on LSUN-Churches while operating 2$\times$ faster with a 3$\times$ smaller architecture.
The project website is available at:
\href{https://toddlerdiffusion.github.io/website/}{$https://toddlerdiffusion.github.io/website/$}
\end{abstract}

%% file: sections/1_Intro.tex
\section{Introduction}
\label{sec_intro}

% -------------------------------------
% Talk about DDPM
In recent years, diffusion models~\cite{sohl2015deep,ho2020denoising,song2020score} have made significant strides across diverse domains, revolutionizing image synthesis and related tasks by transforming noisy, unstructured data into coherent representations through incremental diffusion steps \cite{ho2020denoising, preechakul2022diffusion, dhariwal2021diffusion, nichol2021improved}. 
Their versatility extends beyond image generation to tasks such as image denoising \cite{gong2023pet, kulikov2023sinddm, zhu2023denoising}, inpainting \cite{lugmayr2022repaint, alt2022learning}, super-resolution \cite{li2022srdiff, gao2023implicit}, and applications in 3D content creation \cite{anciukevivcius2023renderdiffusion, qian2023magic123, poole2022dreamfusion}, data augmentation \cite{carlini2023extracting, li2023synthetic, voetman2023big}, medical imaging \cite{chung2022mr, kazerouni2022diffusion, wu2022medsegdiff, wolleb2022diffusion}, anomaly detection, and more.
Diffusion models, such as Denoising Diffusion Probabilistic Models (DDPMs) \cite{ho2020denoising}, draw inspiration from the diffusion process in physics, breaking down the intricate task of generating data from high-dimensional distributions into a sequence of simpler denoising steps. 
In this framework, instead of generating an image in a single step, the task is reformulated as a gradual denoising process, where a noisy image is iteratively refined until the final, detailed image emerges. This step-by-step approach leverages the power of decomposition: by solving a series of simpler tasks, we can ultimately solve the original, more challenging problem.

\input{Figures/1_teaser_fig}

% What: Intro to our apporach
Inspired by this philosophy, we propose {\papernameAbbrev}, a novel method extending this decomposition principle beyond just noise removal. 
Instead of tackling the image generation problem in one go, we decompose it into modality-specific stages. 
Each stage focuses on generating a different aspect or modality of the image, such as abstract contours or sketches, followed by color palettes, and finally, the detailed RGB image. 
This approach mimics the learning process of a child developing artistic skills \cite{stevens2012vision, marr2010vision, marr1974purpose}, where the child first learns to draw basic outlines, then adds colors, and finally fills in the details. 
We use the term ``{\papernameAbbrev}'' to reflect this gradual, step-by-step learning and generation process.

% Challenges:
Despite the intuitive appeal of task decomposition, it introduces significant challenges. 
More stages mean more parameters, longer training times, and potentially slower sampling speeds. 
However, our hypothesis is that by addressing simpler tasks at each stage, we can leverage smaller architectures tailored to each modality, resulting in a more compact overall model compared to traditional single-stage architectures.
Moreover, multi-stage frameworks often suffer from error accumulation \cite{ho2022cascaded, huang2023composer}, where inaccuracies in earlier stages can propagate and amplify through subsequent stages. 
However, we showed that by using simple techniques such as condition truncation dropout, we can significantly mitigate the error accumulation, Appendix \ref{sec_stages_fusion}
Additionally, learning these intermediate representations, e.g., counters, explicitly or implicitly is challenging, where the first requires paired data and the latter requires special complicated designs \cite{singh2019finegan, li2021partgan}.
To tackle the guidance challenge, we leverage pseudo GT and show that this noisy GT is enough to learn robust intermediate modalities.

% How: Excuation way
To implement this framework, we considered two approaches. 
The first is the standard Latent Diffusion Model (LDM) \cite{rombach2022high} conditioning mechanism, using techniques such as cross-attention and concatenation to link stages. 
Given our focus on spatial modalities like sketches, concatenation seemed more appropriate. 
However, this method proved insufficient for achieving our hypothesis due to its weak conditioning, which results in poor information transfer between stages.
We instead adopt a more principled approach by employing the Schr\"odinger Bridge \cite{wang2021deep, li2023bbdm}, a method from stochastic optimal transport theory, to determine the optimal path between modalities at each stage. 
Unlike simple concatenation, which starts each stage from a state of pure noise, the Schr\"odinger Bridge enables us to begin each stage from a meaningful signal with a higher Signal-to-Noise Ratio (SNR) \cite{kingma2021variational}, as illustrated in Figure \ref{fig_trajectory}.
This leads to more effective and stable progression through the generation process, significantly reducing the denoising steps required during training and testing.

% Why: Talk about the gains
This structured, multi-stage approach not only allows for a more efficient and stable generation process but also enables a range of emerging capabilities. {\papernameAbbrev} supports consistent editing of both generated and real images and facilitates user interaction in both conditional and unconditional settings, as shown in Figure \ref{fig_teaser_fic}. 
Moreover, while sampling efficiency was not our primary objective, our approach yields faster sampling and training convergence due to the shorter trajectories between stages (Figure \ref{fig_trajectory}), resulting in fewer denoising steps and less computational cost (Figure \ref{fig_teaser_fic}).
For instance, LDM is trained using 1k steps; in contrast, {\papernameAbbrev} could be trained using only 10 steps with minimal impact on generation fidelity.
To the best of our knowledge, {\papernameAbbrev} is the first work to successfully integrate and achieve advancements across multiple domains, including controllability, consistent editing, sampling and training efficiency, and interactivity, within a unified framework, as shown in Figure \ref{fig_teaser_fic}. 
Our extensive experiments on datasets such as LSUN-Churches \cite{yu2015lsun}, ImageNet, and CelebA-HQ \cite{karras2017progressive} demonstrate the effectiveness of this approach, consistently outperforming existing methods. 
{\papernameAbbrev} not only redefines the image generation process but also paves the way for future research into more structured, interpretable, and efficient generative models.

Our contributions can be succinctly summarized as follows:
\begin{itemize}
\item We introduce a cascaded structured image generation pipeline, denoted as {\papernameAbbrev}, that systematically generates a chain of interpretable stages leading to the final image.
\item Leveraging the Schr\"odinger Bridge enables us to train the model from scratch using a minimal number of steps: 10 steps only, in addition to, performing on the bar compared to LDM while being 2$\times$ faster, using 3$\times$ smaller architecture.
\item {\papernameAbbrev} provides robust editing and interaction capabilities for free; without requiring any manual annotations.
\item Introduce a novel formulation for sketch generation that achieves better performance than LDM while using 41$\times$ smaller architecture.
\end{itemize}

%% file: Figures/1_teaser_fig.tex
% --------------------------------------------------------
\begin{figure*}[t!]
\captionsetup{font=small}
\centering
\includegraphics[width=0.99\linewidth]{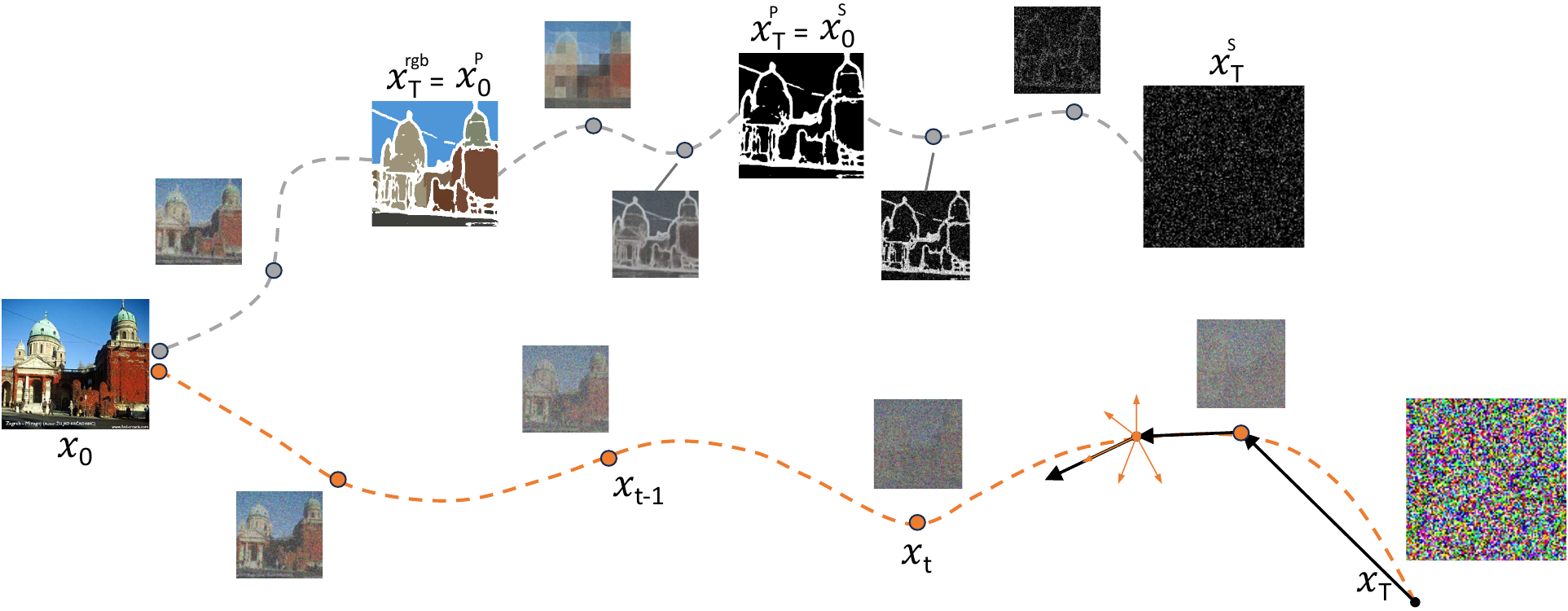}
\caption{Trajectory comparison. We compare our approach (Top) against LDM (bottom) in terms of trajectory. Instead of starting from pure noise such as LDM, our approach, {\papernameAbbrev}, leverages the intermediate stages to achieve a more steady and shorter trajectory.}
\vspace{-3mm}
\label{fig_trajectory}
\end{figure*}
% --------------------------------------------------------

%% file: sections/4_methodology.tex
% ----------------------------
\section{\papernameAbbrev}
\label{sec_methodology}

We introduce {\papernameAbbrev}, which strategically decomposes the RGB image generation task into more straightforward and manageable subtasks, such as sketch/edge and palette modalities. 
In Section \ref{sec_toddler_growth}, we detail the formulation of each stage. By breaking down the generation process into simpler components, our model not only allows for a more compact architecture but also enables dynamic user interaction with the system, as illustrated in Figure \ref{fig_teaser_fic}. 
Notably, these advancements are achieved without relying on manual ground truth annotations, instead utilizing human-free guidance for enhanced efficiency and practical applicability (Section \ref{sec_toddler_guidance}). 
Lastly, we demonstrate how our method facilitates robust, consistent edits across stages (Section \ref{sec_toddler_consistency}).

% ===============================================
\subsection{Toddler Growth}
\label{sec_toddler_growth}

The development of our method is inspired by child's growth, where the learning process in their brains is decomposed into stages and developed step-by-step \cite{stevens2012vision, marr2010vision, marr1974purpose}.
By analogy, we decompose the 2D image synthesis into different, more straightforward stages.
We employ the sketch and palette as intermediate stages, resulting three cascaded stages approach.
However, our approach is generic and seamlessly we can add or remove intermediate steps, based on the application needs.
Mainly, the pipeline can be clustered into two main modules:
1) The unconditional module, that generates the first modality, e.g., sketches, based on pure noise.
2) The conditional module, that condition on the previous modality to generate the next intermediate modality, e.g., palette, or the final RGB image.
Both modules can be implemented using Latent Diffusion Models (LDM), where unconditional LDM will start from noise to generate sketches, then by employing the concatenation conditioning mechanism we can condition on our spatial intermediate modalities; sketches and palettes.
We instead adopt a more principled approach by employing the Schr\"odinger Bridge to determine the optimal path between modalities at each stage. 
Unlike simple concatenation, which starts each stage from a state of pure noise, the Schr\"odinger Bridge enables us to begin each stage from a meaningful signal with a higher Signal-to-Noise Ratio (SNR).

% -----------------------------------------------------
\noindent \textbf{Schr\"odinger Bridge.}
The Schr\"odinger Bridge problem arises from an intriguing connection between statistical physics and probability theory. 
At its core, it addresses the task of finding the most likely stochastic process that evolves between two probability distributions over time, $\rho_0$ and $\rho_T$, governed by Brownian motion, while minimizing the control cost.
Mathematically, it could be written as: 
$P^* = \underset{P \in \mathcal{P}}{\text{argmin}} \, \text{KL}(P \, || \, W)$, where $P$ is the probability path that minimizes the Kullback-Leibler (KL) divergence with respect to a reference Brownian motion $W$, and subject to marginal constrains; $P_0=\rho_0$ and $P_T=\rho_T$.  
The process can be described as a stochastic differential equation (SDE):
{\small
\begin{equation}
\label{eq_SDE}
    dX_t = v(t, X_t) \, dt + \sigma \, dW_t,
\end{equation}
}
where $X_t$ represents the state at time $t$, $v(t, X_t)$ is the optimal drift that minimizes the KL divergence and $dW_t$ is the Wiener process (Brownian motion).

% -----------------------------------------------------
\noindent \textbf{Conditional Module.}
By analogy, the two probability distributions $\rho_T$ and $\rho_0$ represent the distributions of features from simpler (sketch) and more complex (RGB) modalities, respectively. 
In this context, the goal of the Schrödinger Bridge is to find the optimal stochastic process that transforms the simpler modality (e.g., sketch) into the more complex one (e.g., RGB image) while minimizing a divergence measure like KL divergence while respecting the boundaries. 
Given a condition $y \in \mathbb{R}^{H \times W \times 3}$, that could be sketch, palette, or any other modality, we define the forward process as an image-to-image translation and formulate the problem using the discrete version of schr{\"o}dinger bridge as follows:
{\small
\begin{equation}
\label{eq_forward_process_colorstage}
    x_t=\alpha_t x^{i}_0 + (1-\alpha_t)y^{i}  + \sigma^2_t \epsilon_t,
    \quad \quad : \sigma^2_t=\alpha_t - \alpha^2_t,
\end{equation}
}
where $i$ is the stage number, $x_0$ and $y$ are the boundaries of the bridge, $\sigma^2_t$ controls the noise term, and $\alpha_t$ blends the two domains.
In the SDE, the term ``$v(t, X_t)$'' is the deterministic drift, while ``$dW_t$'' represents the diffusion term.
In contrast, we follow the DDPM formulation, which works in the discrete space.
Therefore, by analogy, the terms ``$\alpha_t x_0 + (1-\alpha_t)y$'' and ``$\sigma^2_t \epsilon_t$'' represent the drift and diffusion terms, respectively.  
$\sigma^2_t$ represents uncertainties, following a sphere formulation that constructs a bridge between the two domains.
Whereas, at the edge of the bridge, we are fully confident we are in one of the distributions $\rho_T$ and $\rho_0$, thus $\sigma^2_t=0$. 
In contrast, the uncertainties increase while moving away from the edges, reaching their maximum value in the middle between the two domains, forming the bridge shape.
Eq. \ref{eq_forward_process_colorstage} forms the skeleton for $2^{nd}$ and $3^{rd}$ stages, where:
{\scriptsize
\begin{equation}
    \label{eq_y_foreachstage}
    x^{i}_0 =
    \begin{cases}
    x^{p}_0=\mathcal{F}_p(x^{rgb}_0) & : \dot{\imath}=2 \quad(\text{Palette})\\
    x^{rgb}_0 & : \dot{\imath}=3 \quad(\text{Detailed Image})
    \end{cases}
    \quad \quad \quad \quad
    y^{i} =
    \begin{cases}
    x^{s}_0=\mathcal{F}_s(x^{rgb}_0) & : \dot{\imath}=2 \quad(\text{Palette})\\
    x^{p}_0=\mathcal{F}_p(x^{rgb}_0) & : \dot{\imath}=3 \quad(\text{Detailed Image})
    \end{cases}
\end{equation}
}
where $i$ is the stage number, $\mathcal{F}_p$ and $\mathcal{F}_s$ are palette and sketch functions that generate the pseudo-GT, respectively, detailed in Section \ref{sec_toddler_guidance}.
 
% -----------------------------------------------------
\noindent \textbf{Unconditional Module: $1^{st}$ Stage (Sketch).}
This stage aims to generate abstract contours $x^{s}_0 \in \mathbb{R}^{H \times W}$ starting from just pure noise or label/text conditions. 
\input{Figures/tex/exp_wrap_SketchForward}
In both cases, we name this module unconditional as it does not condition on a previous modality because it will generate the first modality, the sketch. 
One possible solution is to utilize the original LDM, which is not aligned with the sketch nature, as shown in Figure \ref{fig_1st_forward_process}, case A.
More specifically, we are starting from a pure noise distribution $\mathbf{x}_{T}$ and want to learn the inverse mapping $q\left(\mathbf{x}_{t-1} \mid \mathbf{x}_t\right)$ to be able to sample real data $x^s_{0}$.
The issue relies upon the vast gap between the two domains: $x_{T}$ and $x_{0}$.
Following \cite{kingma2021variational}, this can be interpreted as signal-to-noise ratio (SNR), where SNR($t$)=$\frac{\alpha_t}{\sigma^{2}_t}$.
Consequently, at the beginning ($t=T$), $\alpha_T$ yields to 0, therefore SNR($T$)=0.
On the contrary, to fill this gap, we propose a unique formulation tailored for sketch generation that adapts Eq. \ref{eq_forward_process_colorstage}.
Specifically, we introduced two changes to Eq.  \ref{eq_forward_process_colorstage}.
First, we replaced the Gaussian distribution $\epsilon_t$ with a discretized version, the Bernoulli distribution, as shown in case B, Figure \ref{fig_1st_forward_process}.
However, this is not optimized enough for the sparsity nature of the sketch, where more than 80\% of the sketch is black pixels.
Thus, we set our initial condition to $y=\mathcal{F}_{d}(x^{s}_0)$, where $\mathcal{F}_{d}$ is a dropout function that takes the GT sketch $x^{s}_0$ and generates more sparse version.
To align our design with the sketch nature, as shown in Figure \ref{fig_1st_forward_process}, part C, we changed $\sigma^2_t$ to follow a linear trend instead of the bridge shape by setting the variance peak at $x_T$. 
We have to make sure to set the variance peak to a small number due to the sparsity of the sketch.
This could be interpreted as adding brighter points on a black canvas to act as control points during the progressive steps ($T \rightarrow 0$) while converging to form the contours, as shown in Figure \ref{fig_1st_forward_process}, part C.
More details about the sparsity level importance are shown in Section \ref{sec_2nd_multimodal_ablation}.

% ===============================================
\subsection{Toddler Guidance}
\label{sec_toddler_guidance}

A crucial factor for the success of our framework, {\papernameAbbrev}, is obtaining accurate and human-free guidance for the intermediate stages. 
Two modules, $\mathcal{F}_s$ and $\mathcal{F}_p$, are employed to generate contours/sketches and palettes, respectively.

\noindent \textbf{Sketch.}
The first module, $x^{s}_0 = \mathcal{F}_s(x^{rgb}_0)$, serves as a sketch or edge predictor, where $x^{rgb}_0 \in \mathbb{R}^{H \times W \times 3}$ is the ground-truth RGB image, and $x^{s}_0 \in \mathbb{R}^{H \times W}$ is the generated sketch. 
For instance, $\mathcal{F}_s$ could be PidiNet \cite{su2021pixel} or Edter \cite{pu2022edter} for sketch generation, or Canny \cite{canny1986computational} and Laplacian \cite{wang2007laplacian} edge detectors.

\noindent \textbf{Palette.}
The palette can be obtained without human intervention by pixelating the ground-truth RGB image 
$x^{p}_0=\mathcal{F}_p(x^{rgb}_0)$,
where $x^{p}_0 \in \mathbb{R}^{H \times W \times 3}$ is the palette image, and $\mathcal{F}_p$ will be the pixelation function.
However, we introduce a more realistic way to have high-quality palettes for free by first generating the segments from the RGB image using FastSAM \cite{zhao2023fast}. Then, a simple color detector module is utilized to get the dominant color, e.g., the median color per segment.
This design choice is further analyzed in Section \ref{sec_2nd_multimodal_ablation}.
More details are mentioned in the Appendix.

% ==============================================
\subsection{Training Objective \& Reverse Process}
\label{sec_TrainingObjectivesReverseProcess}

% --------------------------------------
\noindent \textbf{Training Formulation.}
We adapt the conventional diffusion models' learning objective \cite{ho2020denoising, sohl2015deep}, i.e., Variational Lower Bound (ELBO), to include our new condition $y$, whereas, each marginal distribution has to be conditioned on $y$, as follows:
{\scriptsize
\begin{equation}
\label{ELBO}
    \mathcal{L}_{ELBO} = -\mathbb{E}_q\big(D_{KL}(q(x_T | x^i_0, y^i) || p(x_T | y^i))
     + \sum^T_{t=2}D_{KL}(q(x_{t-1} | x_t, x^i_0, y^i) || p_{\theta}(x_{t-1} | x_t, y^i)) - \log p_{\theta}(x^i_0 | x_1, y^i)\big),
\end{equation}
}
where $p_{\theta}$ is a function approximator intended to predict from $x_t$.
Using Bayes'rule, we can formulate the reverse process as follows:
{\scriptsize
\begin{equation}
\label{eq_reverse_process_ours}
q\left(x_{t-1} \mid x_t, x^i_0, y^i\right)=\mathcal{N}\left(x_{t-1} ; \tilde{\mu}\left(x_t, x^i_0, y^i\right), \tilde{\sigma}^{2}_t \mathbf{I} \right)
\end{equation}
}
{\scriptsize
\begin{equation}
    \Tilde{\mu}_t(x_t, x^i_0, y^i) = \frac{\sigma^2_{t-1}}{\sigma^2_t}\frac{1-\alpha_t}{1-\alpha_{t-1}}x_t  
     + (1-\alpha_{t-1}(1-\frac{\sigma^2_{t-1}(1-\alpha_t)^2}{\sigma^2_{t}(1-\alpha_{t-1})^2}))x^i_0 
     + (\alpha_{t-1} - \alpha_t\frac{1-\alpha_t}{1-\alpha_{t-1}}\frac{\sigma^2_{t-1}}{\sigma^2_t})y^i
\label{eq_mean_ours}
\end{equation}
}
{\scriptsize
\begin{align}
    \Tilde{\sigma}^2_t = \sigma^2_{t-1} - \frac{\sigma^4_{t-1}}{\sigma^2_{t}}\frac{(1-\alpha_t)^2}{(1-\alpha_{t-1})^2} 
\label{eq_variance_ours}
\end{align}
}

The derivation for Eq. \ref{eq_mean_ours} and Eq. \ref{eq_variance_ours} is detailed in Appendix \ref{appendix_Derivations}.
The overall training and sampling pipelines are summarized in Appendix \ref{sec_algorithms} Algorithms \ref{alg_training_pipeline} and \ref{alg_sampling_pipeline}, respectively.

% --------------------------------------
\noindent \textbf{Training Objectives.}
Previous work \cite{rombach2022high} showed that learning directly the noise $\epsilon$ enhances the performance compared to predicting the $x_0$ directly. 
This is intuitive as predicting the difference $\epsilon_t = x_t-x_0$ is easier than predicting the original complicated $x_0$.
However, in our case, this is not valid due to the conditioned domain $y$, which makes predicting the difference much harder, as shown in Eq. \ref{eq_xt-x0}
{\small
\begin{equation}
    \label{eq_xt-x0}
    x_t-x_0 = \alpha_t (y-x_0) + \sigma_t \epsilon_t
\end{equation}
}
Accordingly, we directly predict the $x_0$. Please refer to Section \ref{sec_2nd_multimodal_ablation} for a detailed analysis.

% ======================================
\subsection{Toddler Consistency}
\label{sec_toddler_consistency}

Our novel framework, {\papernameAbbrev}, inherently provides interactivity, allowing users to observe and engage with the intermediate outputs of various modalities. This interactivity is further strengthened by ensuring consistency, a crucial element for effective control and editing capabilities.

\noindent \textbf{Generated Images Editing.}
To ensure consistency during editing, we store the noise $\epsilon_t$ and the intermediate modalities used in generating the initial reference image, creating a memory that will help us achieve outstanding consistency. 
Throughout the editing session, the memory, that was constructed while generating the reference image, remains fixed until the user resets it by generating a new reference image. 
Figure \ref{fig_exp_controlability_qualitative} demonstrates our framework's editing capabilities, highlighting the strong consistency it achieves. 
Additional editing examples can be found in our demo.

% --------------------------------------
\noindent \textbf{Real Images Editing.}
Real image editing is much more challenging as we cannot access intermediate modalities and noise.
Therefore, given a real image $x^{rgb}_0$, first, we run $\mathcal{F}_s(x^{rgb}_0)$ and $\mathcal{F}_p(x^{rgb}_0)$ to get the corresponding sketch $x^{s}_0$ and palette $x^{p}_0$, respectively.
Then, we utilize the recent DDPM inversion technique \cite{huberman2024edit} to store the noise $\epsilon_t$.
More specifically, to get the noise, we have first to run our forward process, Eq. \ref{eq_forward_process_colorstage}, using random noise $\epsilon_t$ and store the corresponding intermediate states $x_t$.
Then, run our reverse process, Eq. \ref{eq_mean_ours}, to try to recover the original image.
During the reverse process, the model will predict $\Tilde{\mu}_t(x_t, x^i_0, y^i)$ using Eq. \ref{eq_mean_ours}. 
Then, our goal is that given the predicted $\Tilde{\mu}_t$, the predicted conditions $y$ and the stored intermediate states $x_{t-1}$, we want to estimate the noise $\epsilon_t$
Following these two steps, we construct our memory, which will be used similarly to edit real images.

%% file: Figures/tex/exp_wrap_SketchForward.tex
% --------------------------------------------------------
\begin{wrapfigure}{r}{0.45\textwidth}
% \vspace{-6mm}
\captionsetup{font=scriptsize}
\begin{center}
        \includegraphics[width=0.9\linewidth]{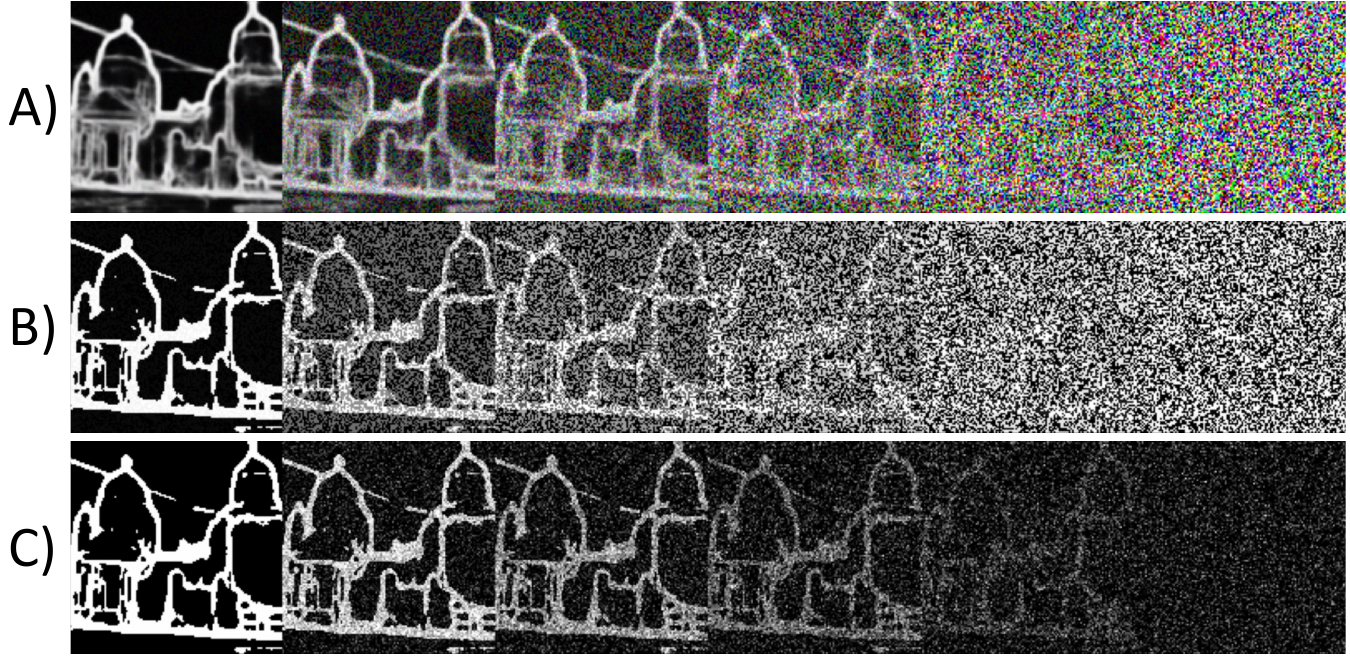}
        \captionof{figure}{Comparison between different formulations for the $1^{st}$ stage. This depicts the forward process for each formulation.}
        \label{fig_1st_forward_process}
\end{center}
\vspace{-6mm}
\end{wrapfigure} 
% ------------------------------------------------------

%% file: sections/5_exp_new.tex
\section{Experimental Results}
\label{sec_experimental_results}

To probe the effectiveness of our proposed framework, we first compare our approach against LDM when trained from scratch on unconditional datasets, LSUN-Churches and CelebHQ, and conditional ones, ImageNet. 
Then, we show that our approach can leverage from the high-quality SD-v1.5 weights, enabling the model to be conditional on new modalities seamlessly without adding additional modules \cite{zhang2023adding, zhao2024uni} or adapters \cite{mou2024t2i}.

% =========================================================
\subsection{{\papernameAbbrev} from Scratch}
\label{sec_exp_ldm_comparision}

In this section, first we will holistically compare our method, {\papernameAbbrev}, against LDM regarding generation quality and efficiency, e.g., convergence rate and number of denoising steps.
Then, we will compare our conditioning ability with ControlNet and SDEdit.
Generally, {\papernameAbbrev} consists of three cascaded stages: sketch, palette, and RGB generation.
However, in the following section, we will show the versatility of our method by omitting the $1^{st}$ or the $2^{nd}$ stages, enabling our method to start from a reference sketch or palette, respectively, instead of generated ones.

% --------------------------------------------------------
\noindent \textbf{Unconditional Generation.}
We evaluated our framework, {\papernameAbbrev}, against LDM on two unconditional datasets, i.e., LSUN-Churches \cite{yu2015lsun} and Celeb datasets \cite{karras2017progressive}.
For a fair comparison, and as we notice challenges in replicating LDM results\footnote{See issues \href{https://github.com/CompVis/latent-diffusion/issues/325}{325}, \href{https://github.com/CompVis/latent-diffusion/issues/262}{262}, \href{https://github.com/CompVis/latent-diffusion/issues/90}{90}, \href{https://github.com/CompVis/latent-diffusion/issues/142}{142}, \href{https://github.com/CompVis/latent-diffusion/issues/30}{30}, and \href{https://github.com/CompVis/latent-diffusion/issues/138}{138} in the LDM GitHub repository.}, we have reproduced LDM results.
As shown in Table \ref{tab_exp_unconditional_benchmark}, we boost the LDM performance by 1 FID point on the two datasets, using 3x smaller architecture and 2x faster sampling, as shown in Table \ref{tab_exp_detailed_complexty}.

% --------------------------------------------------------
\noindent \textbf{Efficient Architecture.}
\input{tables/exp_effecient_arch}
While the primary focus of this paper is not on designing an efficient architecture, our method inherently benefits from architectural efficiency due to the shorter trajectory between stages, as illustrated in Figure \ref{fig_trajectory}. 
The key insight is that {\papernameAbbrev} simplifies each generation task by breaking it into modality-specific stages (e.g., from sketch to palette). 
This approach allows us to leverage significantly smaller networks at each stage because solving the problem is easier, where the distributions of the domains, such as sketches and palettes, are much closer than the distribution between pure noise and RGB images. 
As a result, the neural network at each stage can be more lightweight. Furthermore, as shown in Table \ref{tab_exp_detailed_complexty}, we operate directly in the image space for the first two stages (i.e., noise to sketch and sketch to palette). 
This eliminates the need for complex encoder-decoder architectures, such as VQGAN, typically used to generate detailed images from latent representations. 
Combining these factors, {\papernameAbbrev} achieves a 3x reduction in network size despite cascading across three stages.

% --------------------------------------------------------
\noindent \textbf{Faster Convergence.}
\input{Figures/tex/exp_SamplingSteps}
In addition to outperforming LDM across multiple datasets and metrics, as shown in Table \ref{tab_exp_unconditional_benchmark}, {\papernameAbbrev} exhibits a significantly faster convergence rate. 
We analyzed the convergence behavior by reporting the FID score every 50 epochs, as shown in Figure \ref{fig_exp_ConvergenceRate}. 
Remarkably, our method achieves the same performance after just 50 epochs as LDM does after 150 epochs. 
Furthermore, after 150 epochs, {\papernameAbbrev} matches the performance LDM achieves after 500 epochs. This substantial improvement in convergence speed highlights the strength of our approach in efficiently navigating the feature space between stages.
It’s important to note that faster convergence was not an explicit target of our method but rather an emergent property of the pipeline's design. 
Other techniques tailored explicitly for improving convergence speed, e.g., \cite{hang2023efficient}, can be considered orthogonal and complementary to our method.

% --------------------------------------------------------
\noindent \textbf{Trimming Denoising Steps.}
Another unique characteristic of {\papernameAbbrev} is its ability to reduce the number of denoising steps during both training and sampling without significantly impacting performance. 
As shown in Figure \ref{fig_exp_trimming_denoising_training}, our framework maintains consistent and robust results when the number of steps is reduced from 1000 to 100. 
In contrast, the performance of LDM degrades drastically under the same conditions. {\papernameAbbrev} can trim denoising steps while maintaining a good Signal-to-Noise Ratio (SNR), particularly for larger time steps, as further discussed in Section \ref{sec_toddler_growth}.
Additionally, Figure \ref{fig_exp_trimming_denoising_sampling} highlights another intriguing capability of our method: models trained with fewer denoising steps outperform those trained with larger step counts when later using techniques like DDIM \cite{song2020denoising} to reduce sampling steps. 
For example, as seen in Figure \ref{fig_exp_trimming_denoising_sampling}, the blue curve (trained with 100 steps) achieves an FID score of around 60 when using only ten steps for sampling. 
However, if we initially train the model with just ten steps, as represented by the purple curve, the performance improves significantly, achieving an FID score of 15. 
This demonstrates the efficiency and adaptability of {\papernameAbbrev}, even when minimizing the denoising process.

% --------------------------------------------------------
\noindent \textbf{Class-Label Conditional Generation.}
To holistically evaluate the performance of {\papernameAbbrev}, we tested it on the widely-used ImageNet dataset, where the condition type is a class label. 
Due to limited resources, we conducted our experiments on ImageNet-100, a subset of ImageNet containing 100 classes instead of the full 1,000. 
%While this is a simplified version of the full dataset, we believe this does not undermine the validity of our conclusions. The reduced class count still provides a comprehensive and challenging benchmark to demonstrate the superiority of our approach, as the fundamental complexities of image generation and class-conditional tasks are well preserved.
As shown in Table \ref{tab_exp_unconditional_benchmark}, {\papernameAbbrev} performs better than LDM on key metrics such as FID, KID, Precision, and Recall. 

% --------------------------------------------------------
\noindent \textbf{Controllability.}
\input{Figures/tex/exp_ControlScratch_and_SletchRobust}
We evaluated our editing capabilities against two notable methods: SDEdit \cite{meng2021sdedit} and ControlNet \cite{zhang2023adding}. As illustrated in Figure \ref{fig_ours_vs_sdedit}, SDEdit struggles when conditioned on binary sketches, which is expected given that it is a training-free approach designed to operate with conditions closer to the target domain (i.e., RGB images).
Consequently, compared to ControlNet, a training-based method that adds an additional copy of the encoder network, our approach demonstrates superior faithfulness to the sketch condition, consistency across the generated images, and overall realism. 
Despite ControlNet’s larger model size, our method achieves these improvements without adding any additional parameters. 
Quantitatively, we achieved an FID score of 7, outperforming both SDEdit (15) and ControlNet, which achieved 15 and 9, respectively.

% =========================================================
\subsection{Robustness Analysis}
\label{sec_exp_Robust_Analysis}

\noindent \textbf{Label Robustness.}
Beyond these quality comparisons, we also conducted a robustness analysis to assess our model's behavior when faced with inconsistencies between conditions. 
Specifically, we deliberately provided incorrect class labels for the same sketch.
\input{Figures/tex/exp_wrap_ImagenetRobust}
This mismatched the sketch (geometry) and the desired class label (style). Despite the inconsistent conditions, as illustrated by the off-diagonal images in Figure \ref{fig_exp_ImageNetRobustSubset}, our model consistently adheres to the geometry defined by the sketch while attempting to match the style of the misassigned category, demonstrating the robustness of our approach. Successful cases are highlighted in yellow, and even in failure cases (highlighted in red), our model avoids catastrophic errors, never hallucinating by completely ignoring either the geometry or the style.
More examples can be seen in Appendix \ref{sec_appendix_imgnet_robust}.

% --------------------------------------------------------
\noindent \textbf{Sketch Robustness.}
To further validate the robustness of {\papernameAbbrev}, we designed an aggressive robustness test aimed at evaluating the model's ability to handle sketch perturbations. 
Specifically, we generated three variants of the input sketches using the Structured Edge (SE) detector \cite{dollar2014fast}, varying the kernel size $K$ to introduce different detail and noise levels. 
As shown in Figure \ref{fig_sketch_robustness}, our model consistently demonstrates strong robustness despite these perturbations being absent during training. 
The left part of Figure \ref{fig_sketch_robustness} depicts our model's output using the $2^{nd}$ stage only (Sketch to RGB), while the right part demonstrates the output when using both the $2^{nd}$ and $3^{rd}$ stages.
Even with significant variations in the sketches, {\papernameAbbrev} maintains high fidelity to both the geometry and style, underscoring its ability to adapt to unseen input distortions without compromising output quality.

% =========================================================
\subsection{Editing Capabilities}
\label{sec_exp_Editing}
\input{Figures/tex/exp_celeb_lsun_editing}

As discussed in Section \ref{sec_toddler_consistency}, our approach provides consistent editing capabilities for both generated and real images.

\noindent \textbf{Editing Generated Images.}
In Figure \ref{fig_exp_controlability_qualitative}, we showcase the editing capabilities starting from a generated sketch and RGB image (A). Our method allows for the removal of artifacts or unwanted elements, highlighted in red (B), the addition of new content, shown in yellow (C-D), and modifications to existing content, marked in green (E-F), simply by manipulating the underlying sketch. 
These consistent edits are supported in unconditional and conditional generation scenarios. While sketch-based conditional generation naturally lends itself to such control, ToddlerDiffusion extends this capability to unconditional settings and other conditional models like Text-to-Image. In these cases, we refer to this property as "interactivity," allowing users to seamlessly edit through intermediate modalities (sketch and palette) regardless of the original input format, whether class-label or text.

\noindent \textbf{Editing Real Images.}
As explained in Section \ref{sec_toddler_consistency}, we ensure consistent edits for real images by storing the noise used during the generation process for later edits. The primary challenge with real image editing is that we don’t have access to the noise values at each denoising step, as our model didn’t generate the image. To address this, we employ DDPM-inversion \cite{huberman2024edit} to reconstruct the intermediate noise. 
As illustrated in Figure \ref{fig_teaser_fic}, our model achieves consistent edits on real reference images.
In Appendix \ref{sec_appendix_rela_editing_SEEDx}, we compare against SEED-X \cite{ge2024seed}, and our model achieves more consistent edits, as shown.

% =========================================================
\subsection{Incorporating {\papernameAbbrev} into Stable Diffusion}
\label{sec_exp_SD}

In this section, we will answer an important question: ``Can our method be integrated seamlessly into the existing powerful Stable-Diffusion architecture?
This question is crucial as follows:
1) Despite our approach's great capabilities and performance, leveraging the available powerful weights trained on large-scale datasets such as SD is still important.
2) Our approach follows different forward and reverse processes, Eq. \ref{eq_forward_process_colorstage} and Eq. \ref{eq_mean_ours}, respectively, which could make it challenging to leverage weights that are trained using different training objectives.

\noindent \textbf{Integration Recipe.}
This section will detail the recipe for successfully integrating SD weights into our approach.
First, we used $x_0$ as our default training objective and employed LoRA fine-tuning for efficiency.
But these two variants are crucial to make it work.
The original SD is trained to predict the difference, not $x_0$. 
Thus, we must first try to match its conventional training objective and replace ours by predicting the difference.
However, this will not solve the gap, as our new training objective still includes $x_0$ prediction.
Due to this considerable gap in our formulation compared to SD, we must fine-tune the entire model instead of using LoRA, using a small learning rate, e.g., 1e-6.

\noindent \textbf{Controllability.}
We convert our method to a ``sketch2RGB'' by omitting the $1^{st}$ stage to show our controllability capabilities.
By doing this, we can compare our model against the vast tailored sketch-based models built on top of SD-v1.5.
However, these methods have been trained for a long time on large-scale datasets.
Thus, for a fair comparison, we retrain them using CelebHQ datasets for 5 epochs starting from SD-v1.5 weights.
As demonstrated in Appendix \ref{sec_appendix_Sketch Conditioning(SD-v1.5)}, despite this is not our main focus, our method outperforms conditioning conditioning methods without adding additional modules \cite{zhang2023adding, zhao2024uni} or adapters \cite{mou2024t2i}.

% =========================================================
\subsection{Ablations}
\label{sec_2nd_multimodal_ablation}
\input{Figures/8_ablation_2ndstage_inputs}

% --------------------------------------------------------
\noindent \textbf{$1^{st}$ Stage Representation.}
In Section \ref{sec_toddler_guidance}, we explore the versatility of the $3^{\text{rd}}$ stage (detailed image) by examining six input modalities, detailed in Figure \ref{fig_2nd_multimodal_ablation}. Comparing different contours representations, namely Edges (using Laplacian edge detector \cite{wang2007laplacian}), Sketch (utilizing PidiNet \cite{su2021pixel}), and SAM-Edges (generated by SAM followed by Laplacian edge detector), we find that Sketch outperforms Edges, as edges tend to be noisier. 
However, SAM-Edges provides more detailed contours, yielding superior results. 
Notably, feeding SAM-Colored leads to significant performance degradation, likely due to color discrepancies, as observed in SAM in Figure \ref{fig_2nd_multimodal_ablation}.
While SAM-Edges achieves optimal results, its computational intensity renders it impractical. 
In contrast, Sketch and Edges are computationally inexpensive. 
Furthermore, Sketch's sparse and user-friendly nature makes it more suitable for editing, facilitating interpretation and modification than dense, noisy edges. 
Consequently, we adopt Sketch as the input modality for subsequent experiments.

% --------------------------------------------------------
\noindent \textbf{Palette Effect.}
Adding more guidance will offer more editing abilities to the pipeline and enhance the performance.
As shown in Figure \ref{fig_2nd_multimodal_ablation}, rows b and d, when we incorporate the palette into the contours, i.e., edges and sketch, the performance improved by almost 1 and 1.5, respectively.
In addition, Palette-2 outperforms Palette-1 by 1 FID score.
A more detailed analysis of fusing the stages together and how to mitigate the error accumulation can be found in Appendix \ref{sec_stages_fusion}. 

% --------------------------------------------------------
\noindent \textbf{Concatenation $\textit{v.s.}$ Schr{\"o}dinger Bridge.}
We formulate the generation problem as image-to-image translation. 
For instance, generating a palette given sketch or generating the final RGB image given overlaid palette and sketch.
One possible approach is to utilize the conditioning mechanisms offered by LDM \cite{rombach2022high}, e.g., cross-attention and concatenation.
However, these mechanisms are inefficient as their SNR tends to 0 as $t\!\rightarrow\!T$.
In contrast, our formulation follows the Schr{\"o}dinger bridge, which directly incorporates the condition $y$ into $x_t$ during the forward process. 
Accordingly, the SNR$\gg\!0$ at $t=T$, which allows reducing training steps from 1K to 50 (Figure \ref{fig_exp_trimming_denoising_training}) and achieving faster convergence (Figure. \ref{fig_exp_ConvergenceRate}).
Additionally, Table \ref{tab_cond_mechanisms} demonstrates a quantitative comparison between Schr{\"o}dinger Bridge and concatenation.
As shown in Table, \ref{tab_cond_mechanisms}, the schr{\"o}dinger bridge (SB) combined with the concatenation mechanism achieves the best results; however, SB alone is better than using the naive concatenation.

% --------------------------------------------------------
\noindent \textbf{Training Objectives.}
Previous work \cite{rombach2022high} showed that learning directly the noise $\epsilon$ enhances the performance compared to predicting the $x_0$. 
As, intuitively, predicting the difference $\epsilon_t=x_t-x_0$ is easier than predicting the original complicated $x_0$.
However, this is not valid in our case due to the new term $y$, which makes predicting the difference much harder, as discussed in Section \ref{sec_toddler_growth}.
Table \ref{tab_trainingobjectives} shows the same conclusion reported in \cite{rombach2022high}, that predicting the difference $\epsilon$ leads to better performance than $x_0$, where the FID score is 8.7 and 15.1, respectively.
On the contrary, our method achieves almost the same performance regardless of the training objective, which is expected, as predicting the difference involves predicting $x_0$, Eq. \ref{eq_xt-x0}.

%% file: tables/exp_effecient_arch.tex
% --------------------------------------------------------
\begin{table}[t!]
    \centering
    \captionsetup{font=scriptsize}
    \begin{minipage}{.44\linewidth}
        \begin{center}
            \captionof{table}{Benchmark results on three datasets; CelebHQ \cite{karras2017progressive}, LSUN-Churches \cite{yu2015lsun} and ImageNet \cite{deng2009imagenet}.}
            \label{tab_exp_unconditional_benchmark}            
            \scalebox{0.6}{
            \begin{tabular}{ccccccc}
                    \toprule
                     \textbf{Dataset} & 
                     \textbf{Method} &
                     \textbf{Epochs}  & 
                     \textbf{FID} $\downarrow$ & 
                     \textbf{KID} $\downarrow$ &
                     \textbf{Prec.} $\uparrow$ &
                     \textbf{Recall} $\uparrow$ \\
                     \cmidrule(r){1-1}
                     \cmidrule(r){2-2}
                     \cmidrule(r){3-3}
                     \cmidrule(r){4-7}
                     \multirow{2}{*}{{\textbf{CelebHQ}}} & LDM  & 600 & 8.15 & 0.013 & 0.52 & 0.41 \\
                     \- &  Ours & 600 & \textbf{7.10}  & \textbf{0.009} & \textbf{0.61} & \textbf{0.47}  \\
                     \midrule
                     \multirow{2}{*}{{\textbf{Churches}}} & LDM  & 250 & 7.30 & 0.009 & 0.59 & 0.39 \\
                     \- &  Ours & 250 & \textbf{6.19}  & \textbf{0.005} & \textbf{0.71} & \textbf{0.44}  \\
                     \midrule
                     \multirow{2}{*}{{\textbf{ImageNet}}} & LDM  & 350 & 8.55 & 0.015 & 0.51 & 0.32 \\
                     \- &  Ours & 350 & \textbf{7.8}  & \textbf{0.010} & \textbf{0.58} & \textbf{0.40}  \\
                    \bottomrule
                    \end{tabular}
            }
        \end{center}
    \end{minipage}
    \hfill
    \begin{minipage}{.53\linewidth}
        \begin{center}
            \captionof{table}{A detailed analysis of our method complexities compared to LDM. We report the training and sampling throughput per each stage. For efficiency, the first two stages are operating on the pixel space. Therefore, the VQGAN encoder and decoder are not used (N/A). The training time is reported till convergence, i.e, 400 epochs, using 4 A100.}
            \vspace{-3mm}
            \label{tab_exp_detailed_complexty}            
            \scalebox{0.57}{
            \begin{tabular}{ccccc}
                \toprule
                 &  \textbf{VQGAN} & \textbf{UNet} & \textbf{Train} & \textbf{Samp.} \\
                \textbf{Stage} & \# \textbf{Param} & \# \textbf{Param} & \textbf{Time} & \textbf{Time} \\ \midrule
                1) Noise $\rightarrow$ Sketch & N/A & 1.9M & 1\,hr& 85\,fps\\
                2) Sketch $\rightarrow$ Palette & N/A & 1.9M & 1\,hr& 85\,fps\\
                3) Palette $\rightarrow$ RGB & 55.3M & 101M & 24\,hr& 2.3\,fps\\
                Ours (3 Stages) & 55.3M & \textbf{104.8M} & \textbf{26hr}& \textbf{2.27fps}\\
                LDM & 55.3M & 263M& 38\,hr & 1.21\,fps\\
                \bottomrule
            \end{tabular}
            }
        \end{center}
    \end{minipage}
\end{table}
% --------------------------------------------------------

%% file: Figures/tex/exp_SamplingSteps.tex
%%%%%%%%%%%%%%%%%%%%%%%%    Figure      %%%%%%%%%%%%%%%%%%%%
\begin{figure*}[t!]
    \captionsetup{font=scriptsize}
    \centering
        \begin{minipage}{.3\linewidth}
            \includegraphics[width=\textwidth]{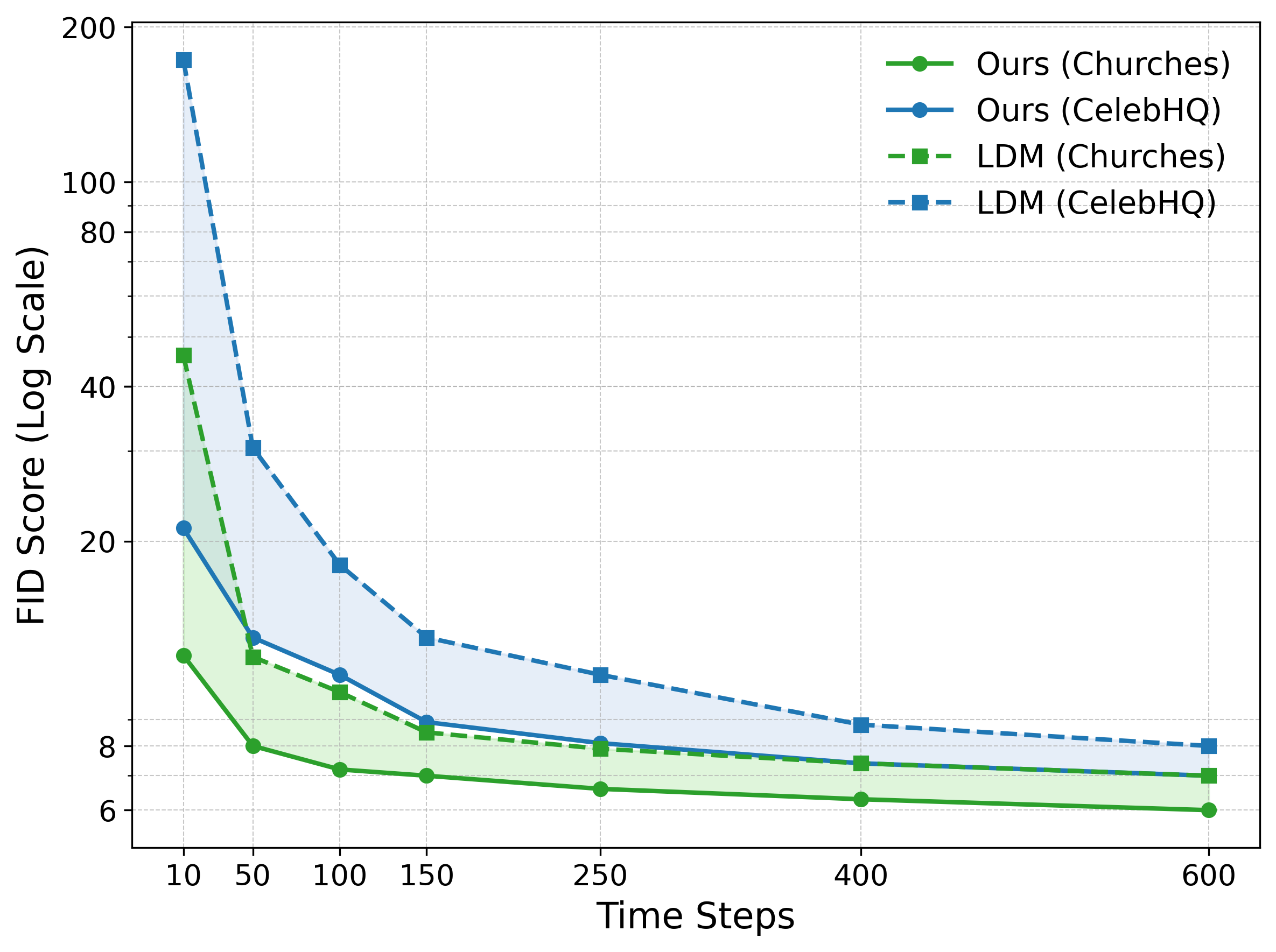}
            \captionof{figure}{Training convergence comparison between {\papernameAbbrev} and LDM.}
            %\vspace{-1.0em}
            \label{fig_exp_ConvergenceRate}
        \end{minipage}
        \hfill
        \begin{minipage}{.3\linewidth}
            \includegraphics[width=\textwidth]{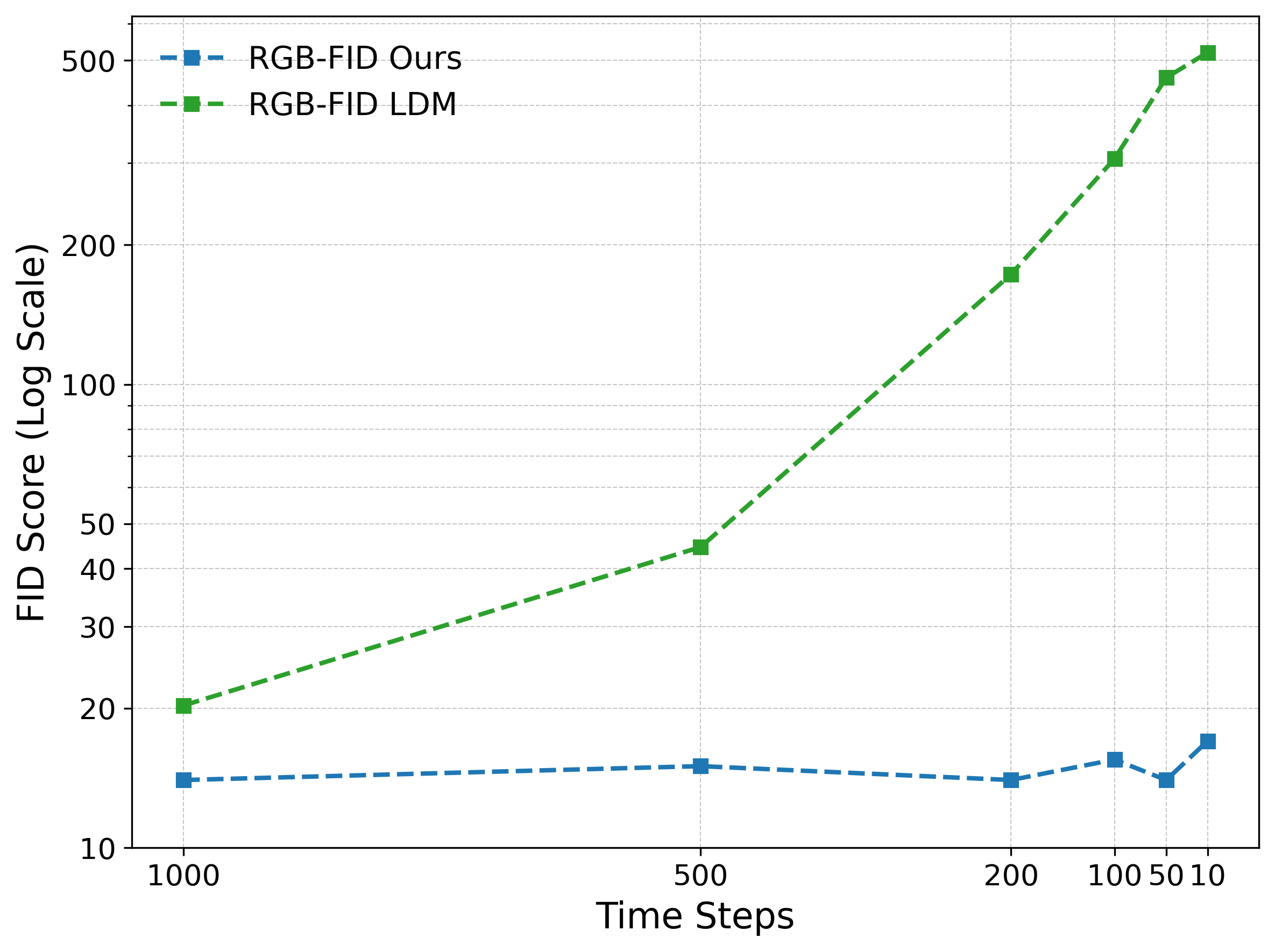}
            \captionof{figure}{Trimming denoising steps ablation study during training and sampling.}
            %\vspace{-1.0em}
            \label{fig_exp_trimming_denoising_training}
        \end{minipage}
        \hfill
        \begin{minipage}{.3\linewidth}
            \includegraphics[width=\textwidth]{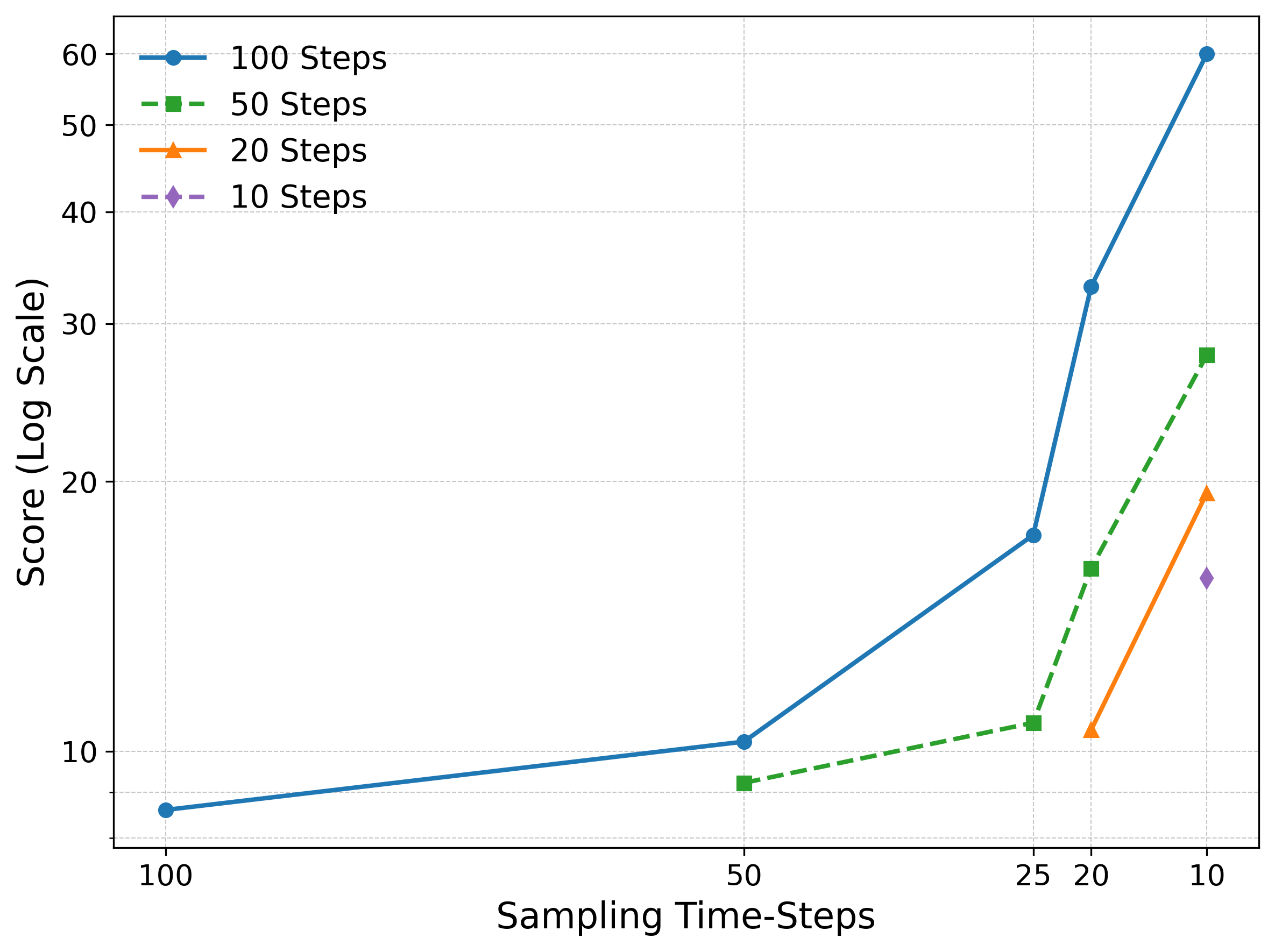}
            \captionof{figure}{Training steps ablation study on LSUN-Churches dataset.}
            %\vspace{-1.0em}
            \label{fig_exp_trimming_denoising_sampling}
        \end{minipage}
    \vspace{-1em}
\end{figure*}
%%%%%%%%%%%%%%%%%%%%%%%%    Figure      %%%%%%%%%%%%%%%%%%%%

%% file: Figures/tex/exp_ControlScratch_and_SletchRobust.tex
%%%%%%%%%%%%%%%%%%%%%%%%    Figure      %%%%%%%%%%%%%%%%%%%%
\begin{figure}[t!]
    \captionsetup{font=scriptsize}
    \centering
        \begin{minipage}{.27\linewidth}
            \includegraphics[width=1\textwidth]{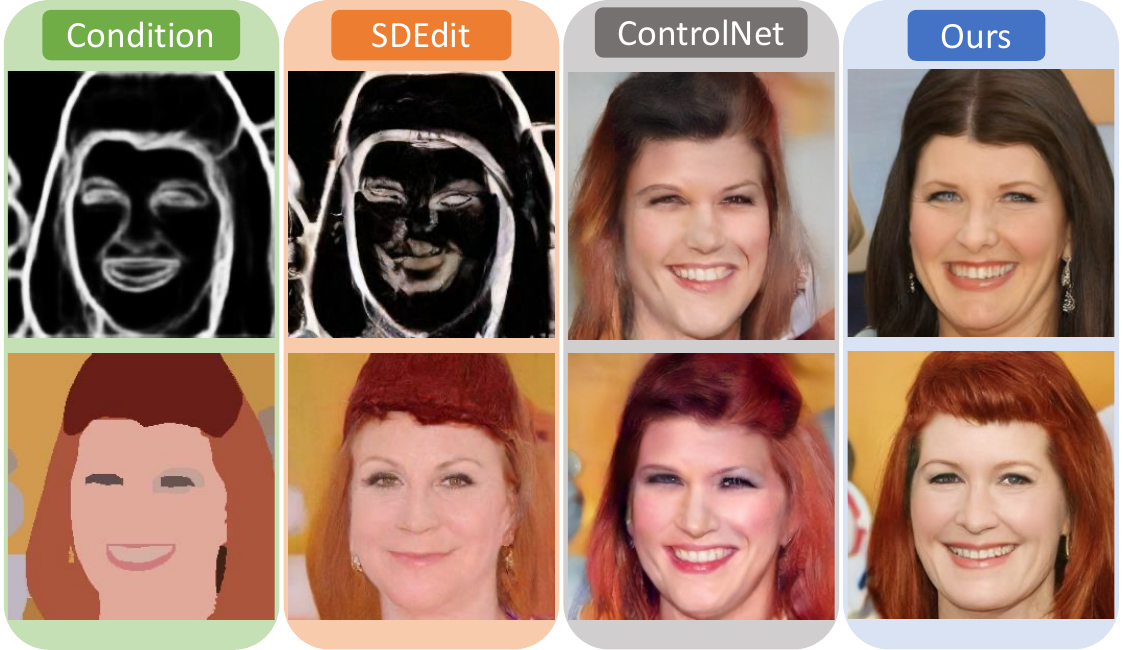}
            \vspace{-6.0mm}
            \captionof{figure}{Comparison between our editing capabilities and SDEdit \cite{meng2021sdedit}.}
            \label{fig_ours_vs_sdedit}
        \end{minipage}
        \hfill
        \begin{minipage}{.7\linewidth}
            \includegraphics[width=0.99\textwidth]{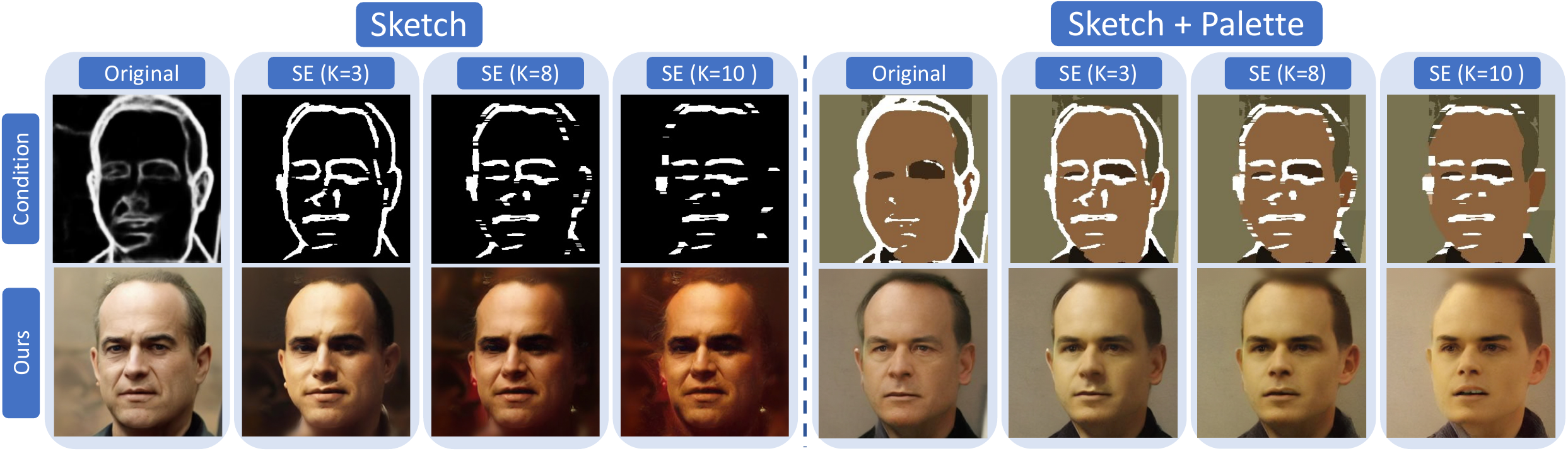}
            \vspace{-3.0mm}
            \captionof{figure}{Sketch robustness analysis.}
            %\vspace{-1.0em}
            \label{fig_sketch_robustness}
        \end{minipage}
        \vspace{-1.0em}
\end{figure}
%%%%%%%%%%%%%%%%%%%%%%%%    Figure      %%%%%%%%%%%%%%%%%%%%

%% file: Figures/tex/exp_wrap_ImagenetRobust.tex
% --------------------------------------------------------
\begin{wrapfigure}{r}{0.43\textwidth}
\centering
% \vspace{-6mm}
\captionsetup{font=scriptsize}
% \captionsetup{font=small}
\includegraphics[width=0.9\linewidth]{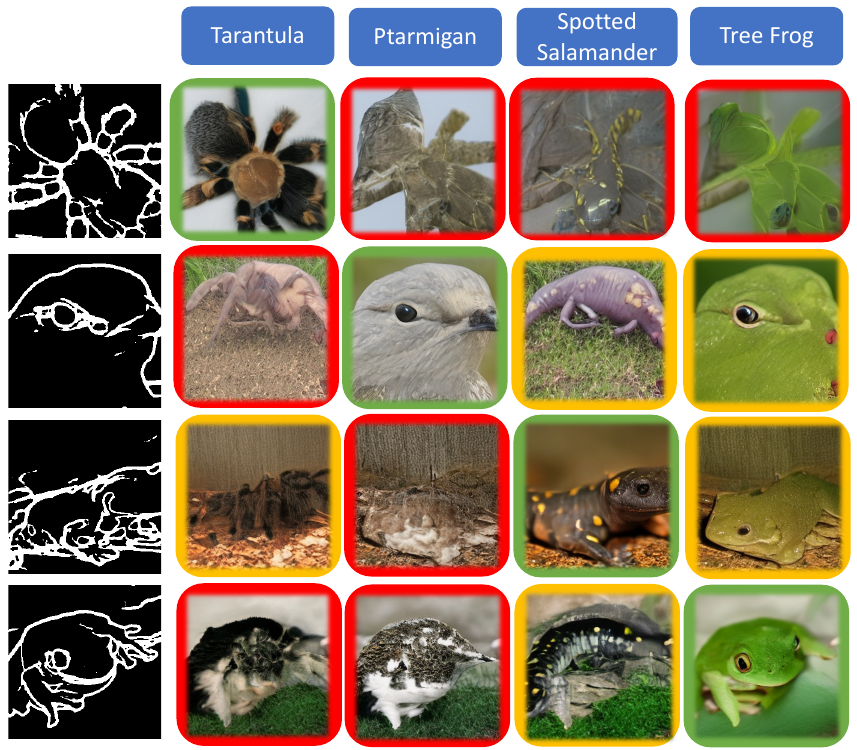}
\caption{Our results on the ImageNet dataset. The first column depicts the input sketch. The diagonal (green) represents the generated output that follows the sketch and the GT label. The off-diagonal images show the output given the sketch and inconsistent label. These adversarial labels show our model's robustness, where the successful cases are in yellow, and the failure ones are in red.}
\vspace{-6mm}
\label{fig_exp_ImageNetRobustSubset}
\end{wrapfigure} 
% ------------------------------------------------------

%% file: Figures/tex/exp_celeb_lsun_editing.tex
% --------------------------------------------------------
\begin{figure*}[t!]
\captionsetup{font=scriptsize}
\centering
\includegraphics[width=0.99\linewidth]{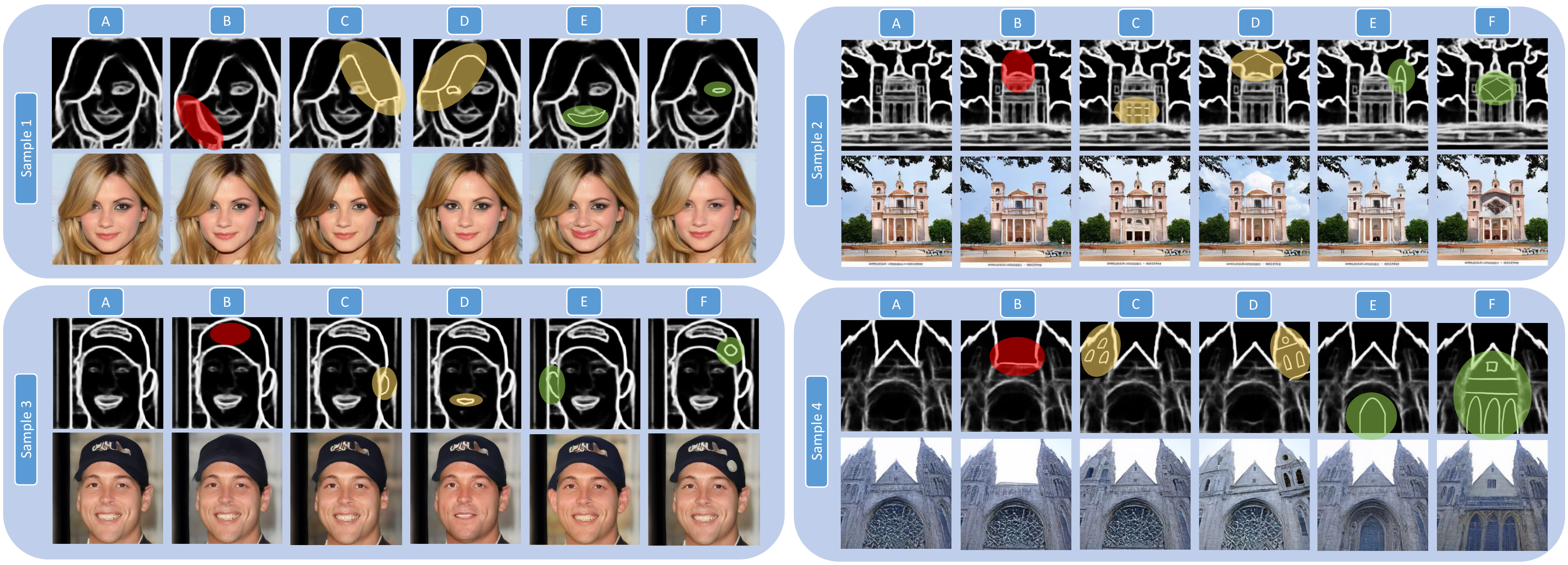}
\caption{Controllability ability of our framework, {\papernameAbbrev}.
Starting from the generated sketch and RGB image (A), we can remove artifacts or undesired parts in red (B), add new content in yellow (C-D), and edit the existing content in green (E-F) by manipulating the sketch.}
\label{fig_exp_controlability_qualitative}
\vspace{-5mm}
\end{figure*}
% --------------------------------------------------------

%% file: Figures/8_ablation_2ndstage_inputs.tex
% --------------------------------------------------------
\begin{figure}[t!]
    \captionsetup{font=scriptsize}
    \begin{minipage}[c]{.45\linewidth}
        \captionof{figure}{Ablation study for different input's types for the $2^{nd}$ stage on LSUN-Churches dataset \cite{yu2015lsun}.}
        \label{fig_2nd_multimodal_ablation}
        \begin{minipage}[c]{.62\linewidth}
            %\vspace{0pt}
            \begin{center}
            \includegraphics[width=0.9\linewidth]{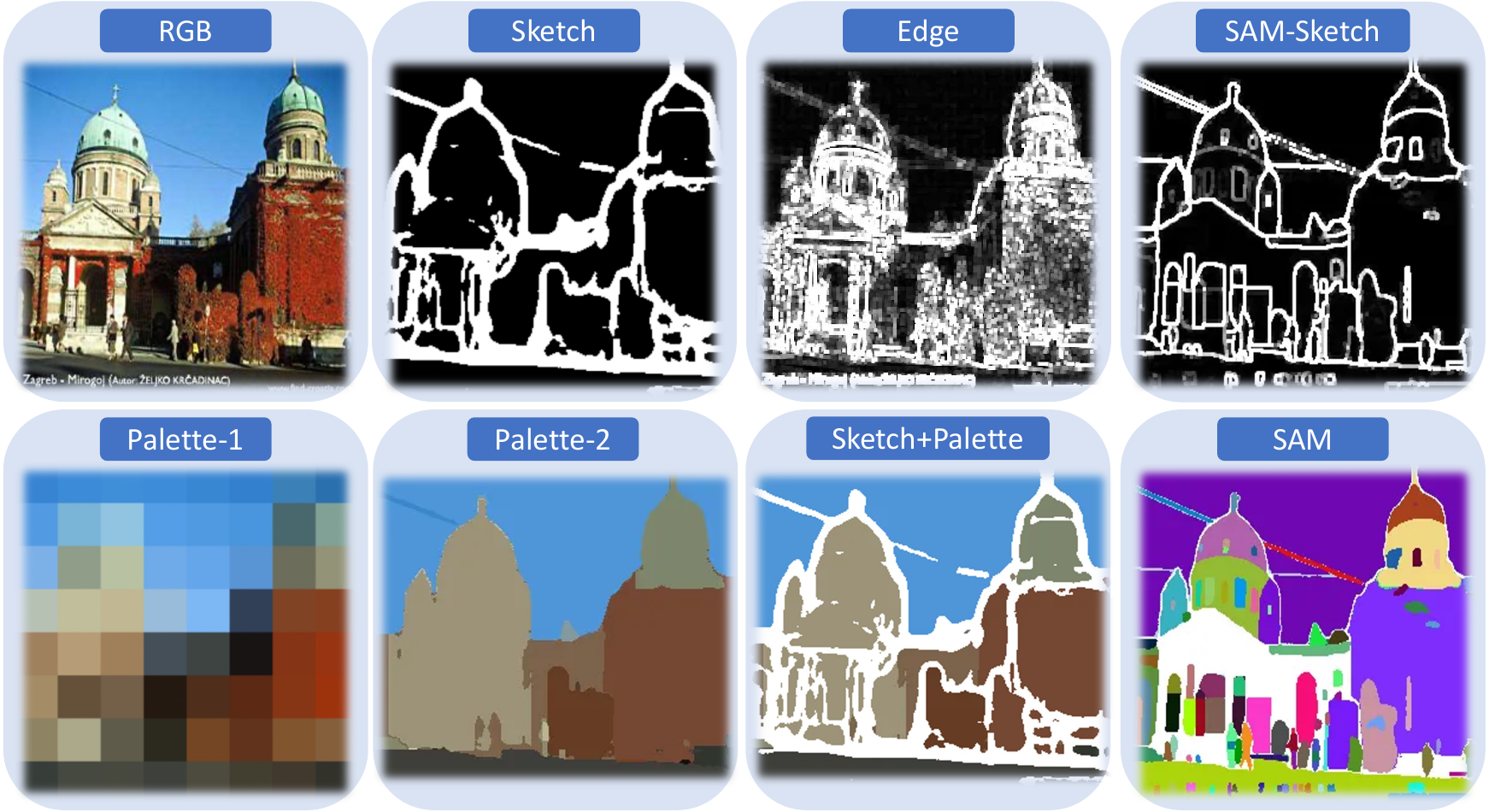}
            \end{center}
        \end{minipage}
        \begin{minipage}[c]{.25\linewidth}
            %\vspace{0pt}
            \begin{center}
            %\captionof{table}{}
            \scalebox{0.5}{
            \begin{tabular}{ccc}
                \toprule
                \- & \textbf{Input Type} & \textbf{FID} $\downarrow$\\
                \midrule
                a) & Edges            & 9.5  \\
                b) & Edges + Palette-2  & 8.4  \\
                c) & Sketch           & 8.6  \\
                d) & Sketch + Palette-1 & 8.3  \\
                e) & Sketch + Palette-2 & \textbf{7.1}  \\
                f) & SAM (Colored)    & 13.1 \\
                g) & SAM (Edges)      & 8.1  \\
                \bottomrule
                \end{tabular}
            }
            \end{center}
        \end{minipage}
    \end{minipage}
    \hfill
    \hfill
    \begin{minipage}{.2\linewidth}
        \begin{center}
            \captionof{table}{Comparison of the LDM concatenation mechanism against the schr{\"o}dinger bridge (SB) using CelebHQ dataset.}
            \vspace{-1mm}
            \label{tab_cond_mechanisms}            
            \scalebox{0.55}{
            \begin{tabular}{cc}
                    \toprule
                     $2^{nd}$ \textbf{Stage} &  \textbf{FID} $\downarrow$ \\
                     \cmidrule(r){1-1}
                     \cmidrule(r){2-2}
                     Concat & 10.04 \\
                     SB &  9.45 \\
                     SB + Concat & \textbf{8.10} \\
                    \bottomrule
                    \end{tabular}
            }
            \end{center}
    \end{minipage}
    \hfill
    \begin{minipage}{.28\linewidth}
        \begin{center}
            \captionof{table}{Training objective ablation study on CelebHQ dataset \cite{karras2017progressive}.}
            \label{tab_trainingobjectives}            
            \scalebox{0.6}{
            \begin{tabular}{ccccc}
                    \toprule
                     \multicolumn{1}{c}{\multirow{2}{*}{{\textbf{Epochs}}}}  
                     &  \multicolumn{2}{c}{\textbf{LDM}}  & \multicolumn{2}{c}{\textbf{Ours}} \\
                     \- &  $(x_t-x_0)$ & $x_0$ & ($x_t-x_0)$ & $x_0$ \\
                     \cmidrule(r){1-1}
                     \cmidrule(r){2-3}
                     \cmidrule(r){4-5}
                     50  & 12.7 & 15.7 & 8.5 & 8.1 \\
                     100 & 10.0 & 14.6  & 8.4 & 7.8 \\
                     200 & 8.7  & 15.1 & 8.2 & \textbf{7.7}\\
                    \bottomrule
                    \end{tabular}
            }
            \end{center}
    \end{minipage}
    % \vspace{-6mm}
\end{figure}
% --------------------------------------------------------

%% file: sections/7_conclusion.tex
\section{Conclusion}

{\papernameAbbrev} introduces a paradigm shift in image generation by breaking down the complex task into more manageable, modality-specific stages. 
Inspired by the natural learning process of a child, our framework elegantly mirrors how human artists progress from rough sketches to fully detailed images. 
This structured approach allows us to simplify each stage of the generation process, enabling the use of smaller architectures and achieving efficient, high-quality results.
Despite the potential challenges introduced by multi-stage frameworks, such as increased parameter counts and error accumulation, {\papernameAbbrev} overcomes these obstacles through the strategic use of the Schrödinger Bridge, which ensures smooth transitions between modalities. 
By maintaining a high Signal-to-Noise Ratio (SNR) at each stage, we enable stable, effective progression, minimizing the number of denoising steps required during both training and sampling. 
This results in faster convergence, lower computational cost, and enhanced sampling efficiency.
Moreover, {\papernameAbbrev} provides a level of control and interaction that is unprecedented in current diffusion-based models, offering consistent editing capabilities across both generated and real images.
Through extensive experiments on multiple datasets, we have shown that {\papernameAbbrev} outperforms existing state-of-the-art models like LDM in terms of quality, interactivity, and speed. 
The flexibility and versatility of our approach make it applicable to a wide range of tasks, and its emergent properties, such as faster convergence and improved editing control, further solidify its potential for real-world applications.

%% file: sections/appendix.tex
\appendix
% ====================================================
\section{{\papernameAbbrev} Method (More Details)}
\label{sec_appendix_method}

% -----------------------------------------
\subsection{Derivations}
\label{appendix_Derivations}

\noindent
\begin{table}[ht!]
\captionsetup{font=scriptsize}
\centering
\label{tab:derivations}
\begin{minipage}[c]{0.95\linewidth}
    \noindent\rule{\linewidth}{0.8pt} \\
    \text{From the forward process, we have:}
    \begin{align}
        &x_t = (1- \alpha_t)x_0 +  \alpha _t y + \sigma_t  \epsilon_t \label{eq_xt_forwardprocess} \\
        &x_{t-1} = (1- \alpha_{t-1})x_0 +  \alpha _{t-1} y + \sigma_{t-1}  \epsilon_{t-1} \label{eq_xt-1_forwardprocess}
    \end{align}
    
    \text{From \eqref{eq_xt-1_forwardprocess}, we get:}
    \begin{equation}
        x_0 =  \frac{1}{1- \alpha_{t-1}}x_{t-1} -  \frac{ \alpha_{t-1}}{1- \alpha_{t-1}} y - \frac{ \sigma_{t-1}}{1- \alpha_{t-1}} \epsilon_{t-1} 
        \label{eq_x0_given_xt-1_from_forwardprocess}
    \end{equation}

    \text{Inserting \eqref{eq_x0_given_xt-1_from_forwardprocess} into \eqref{eq_xt_forwardprocess}, we get:}
    \begin{align}
        q(x_t \mid x_{t-1},x_0) =  \frac{1- \alpha_t}{1- \alpha_{t-1}} x_{t-1} - \frac{(1- \alpha_t) \alpha_{t-1}}{1- \alpha_{t-1}} y  \nonumber \\
        + \alpha_t y - \frac{(1- \alpha_t)  \sigma_{t-1}}{1- \alpha_{t-1}}  \epsilon_{t-1} + \sigma_t \epsilon_t 
        \label{eq_q_xt_xt-qandx0}
    \end{align}

    \text{The reverse process follows:}
    \begin{equation}
        q\left(x_{t-1} \mid x_t, x_0, y\right)=\mathcal{N}\left(x_{t-1} ; \tilde{\mu}\left(x_t, x_0, y\right), \tilde{\sigma}^{2}_t \mathbf{I} \right)
        \label{eq_reverse_process}
    \end{equation}

    \text{Using Bayes' rule:}
    \begin{equation}
        q\left(x_{t-1} \mid x_t, x_0, y\right)= q(x_t \mid x_{t-1} , x_0) \frac{q(x_{t-1}  \mid  x_0)}{q(x_t  \mid  x_0)} 
        \label{eq_reverse_process_Bayes'Rule}
    \end{equation}

    \text{Plugging \eqref{eq_xt_forwardprocess}, \eqref{eq_xt-1_forwardprocess}, and \eqref{eq_q_xt_xt-qandx0} into \eqref{eq_reverse_process_Bayes'Rule}, yields:}
    \begin{align}
        \Tilde{\mu}_t(x_t, x_0, y) =  \frac{\sigma^2_{t-1}}{\sigma^2_{t}} \frac{1- \alpha_t}{1-\alpha_{t-1}} x_t + (1- \alpha_{t-1} \left( 1 - \frac{\sigma^2_{t-1}}{ \sigma^2_t} \frac{(1- \alpha_t)^2 }{ (1- \alpha_{t-1})^2} \right)) x_0 \nonumber \\
        + (\alpha_{t-1} - \alpha_t \frac{(1- \alpha_t)}{(1- \alpha_{t-1})} \frac{ \sigma^2_{t-1}}{\sigma^2_{t}}) y 
    \end{align}

    \begin{equation}
        \tilde{\sigma}^2_t = \sigma^2_{t-1} - \frac{\sigma^4_{t-1}}{\sigma^2_{t}}\frac{(1-\alpha_t)^2}{(1-\alpha_{t-1})^2} 
        \label{eq_variance_ours_derivation}
    \end{equation}

    \text{Finally, we can apply the DDIM non-Markovian forward processes.}
    \text{In return, only the $x_0$ coefficient will be affected.} \\
    \noindent\rule{\linewidth}{0.8pt}
\end{minipage}
\end{table}
% -----------------------------------------

% ---------------------------------------------------------------------------

% -----------------------------------------
\subsection{Architecture Details}
\label{sec_appendix_architecture_details}

Figure \ref{fig_cot_arch} depicts the details of our architecture.
We adopt the LDM architecture, which consists of two main blocks: 
1) VQGAN Encoder-Decoder to convert the image into low-cost latent space. 
2) Unet that responsible for predicting $x_{t-1}$.
Figure \ref{fig_cot_arch} shows that the condition $y$ is encoded into $x_t$. However, it could be optionally concatenated to $x_t$.
In addition, shown in orange and green are the $\sigma^2$ and $\alpha$, respectively.

\input{Figures/tex/arch_horizontal}

% -----------------------------------------
\subsection{$1^{st}$ Stage: Abstract Structure}
\label{sec_exp_1ststage}
\input{Figures/7_1ststage_results}
% --------------------------------------------------------
\noindent \textbf{Sketch FID.}
A discrepancy exists between the reported conventional FID (RGB-FID), trained on ImageNet \cite{deng2009imagenet}, and qualitative results, as illustrated in Figure \ref{fig_qualitative_1st}. This discrepancy \cite{ge2020creative} may arise from differences between the training data (RGB images) and the evaluation data (binary images). To bridge this gap, we introduce Sketch-FID by re-training the inception model \cite{szegedy2015going} on a sketch version of the ImageNet dataset. We generate sketches for ImageNet RGB images using PidiNet \cite{su2021pixel} and train the inception model on the conventional classification task.

% --------------------------------------------------------
\noindent \textbf{Noise Scheduler.}
In the $1^{st}$ stage, where our starting point $y$ is a black image (Section \ref{sec_toddler_growth}), designing an appropriate noise scheduler is crucial. The bridge noise scheduler is intuitively unsuitable, as it eliminates randomness by adding no noise at both edges, fixing the starting point to a black image. This hypothesis is supported by empirical results in Figure \ref{fig_qualitative_1st}, row b, where the model outputs random patterns. We explored linear and logarithmic schedulers, finding the linear schedule superior, yielding Sketch-FID scores of 15.19 and 18.47, respectively (Figure \ref{fig_qualitative_1st}, rows c-d).

% --------------------------------------------------------
\noindent \textbf{Ours vs. LDM.}
In Section \ref{sec_toddler_growth}, we propose an alternative way to generate sketches by leveraging LDM \cite{rombach2022high}, as shown in Figure \ref{fig_1st_forward_process}, row A. However, this approach deviates from the nature of sketching. Our proposed formulation, aligned with the topology of sketches (Figure \ref{fig_1st_forward_process}, row C), resulted in significant improvements over LDM in both model complexity and performance, as depicted in Figure \ref{fig_qualitative_1st}. Our formulation (Section \ref{sec_toddler_growth}) allows direct operation on the image space ($64 \times 64$) and compression of the Unet to a tiny variant without sacrificing performance. Despite the aggressive compression, our performance is significantly better than LDM, with respective Sketch-FID scores of 15.19 and 49, using a 41x smaller network.

% --------------------------------------------------------
\noindent \textbf{Sketch Intensity.}
\input{Figures/appendix/paletteFig_and_sketchTable}
Controlling the intensity is crucial. Thus, we conducted further analysis, as shown in Table \ref{tab_intensity_ablation}, that probes for best performance, the intensity should follow the data distribution.
We refer to the intensity as the ratio between the white pixels to the black ones in the sketch.
The reported FID scores in Table \ref{tab_intensity_ablation} are the overall FID using all stages, and the intensity numbers are the initial intensity at $t=T$.
The GT intensity for CelebHQ and Lsun-Churches are 19.2\% and 23.5\%, respectively.

% -----------------------------------------
\subsection{$2^{nd}$ Stage: Palette}
\label{sec_exp_2ndstage}

As depicted in Figure \ref{fig_palette_pipeline}, we introduce a more realistic way to have high-quality palettes for free by first generating the segments from the RGB image using FastSAM \cite{zhao2023fast}. Then, a simple color detector module is utilized to get the dominant color, e.g., the median color per segment.

% -----------------------------------------
\subsection{Training and Sampling Algorithms}
\label{sec_algorithms}

% -----------------------------------------------
\begin{figure}[H]
    \centering
    \scalebox{0.95}{
    \begin{minipage}{.9\textwidth}
        \begin{algorithm}[H]
            {
            \scriptsize{
                \caption{\small Training Pipeline}
                \label{alg_training_pipeline}
                \begin{algorithmic}[1]
                    \For{$i = 1, \dots , N$} \algorithmiccomment{Train each stage separately}
                        \For{$j = 1, \dots , E$} \algorithmiccomment{Train for E epochs}
                        \For{$(x^i_0, y^i) \in \mathcal{D}$} \algorithmiccomment{Loop on the dataset $\mathcal{D}$}
                            %\State $(x_0, y) \in \mathcal{D}$ \algorithmiccomment{Sampling from the data}
                            \State $\epsilon \in \mathcal{N}(0, I)$ \algorithmiccomment{Sampling the noise}
                            \State $t \in \textit{U}(1, \dots, T)$ \algorithmiccomment{Sampling time-step}
                            %\State $\tilde{x}_0 \in \{\mathcal{F}_{d}(x_0, t), \mathcal{F}_p(x_0, \mathcal{K}_t, \mathcal{J}_t), x_0\}$ \algorithmiccomment{$\tilde{x}_0$ based on the stage}
                            \State Get $x^i_0$ by applying Eq. \ref{eq_y_foreachstage} \algorithmiccomment{$\tilde{x}_0$ based on the stage}
                            \State $x_t=\alpha_t x^i_0 + (1-\alpha_t) y^i + \sigma^2_t \epsilon_t$ \algorithmiccomment{Forward process}
                            \State $\nabla_\theta\left\| \tilde{x}^i_0-p_{\theta}(x^i_{0} | x_t, y^i) \right\|^2$ \algorithmiccomment{Simplified objective}
                        \EndFor
                        \EndFor
                    \EndFor
                   \end{algorithmic}
                }
            }
            \end{algorithm}
    \end{minipage}
    }
    %\hfill
    \scalebox{0.95}{
    \begin{minipage}{.9\textwidth}
        \begin{algorithm}[H]
            {
            \scriptsize{
                \caption{\small Sampling Pipeline}
                \label{alg_sampling_pipeline}
                \begin{algorithmic}[1]
                \State $x_{inter} = []$ \algorithmiccomment{Initialize the list to save intermediate output for each stage}
                \State $y^1 = Zeros((H, W, 1))$ \algorithmiccomment{Initialize $1^{st}$ condition as a black image}
                \State $\epsilon \in \mathcal{N}(0, I)$ \algorithmiccomment{Sampling the noise once}
                    \For{$i = 1, \dots , N$} \algorithmiccomment{Sequential Sampling}
                        %\State $\tilde{x}_0 \in \{\mathcal{F}_{d}(x_0, t), \mathcal{F}_p(x_0, \mathcal{K}_t, \mathcal{J}_t), x_0\}$ \algorithmiccomment{$\tilde{x}_0$ based on the stage}
                        \State $x_T=y^i + \sigma^2_T \epsilon_T$ \algorithmiccomment{Run forward process once}
                        \For{$t = T, \dots , 1$} \algorithmiccomment{Progressive Sampling}
                            \State Get $x_{t-1}$ by applying Eq. \ref{eq_mean_ours} :$x^{i}_0=p_{\theta}(x_{0} | x_t)$ \algorithmiccomment{Reverse process}
                        \EndFor
                        \State $y^{i+1} = x^{i}_0$ \algorithmiccomment{Update the condition for the next stage}
                        \State $x_{inter}.append(x^{i}_0)$ \algorithmiccomment{Store the intermediate outputs for each stage}
                    \EndFor
                    \State \Return $x_{inter}$
                   \end{algorithmic}
                }
            }
        \end{algorithm}
    \end{minipage}
    }
\end{figure}
% -----------------------------------------------

% -----------------------------------------
\subsection{How to Fuse different Stages Efficiently?}
\label{sec_stages_fusion}

%%%%%%%%%%%%%%%%%%%%%%%%    Figure      %%%%%%%%%%%%%%%%%%%%
\begin{figure}[H]
    \captionsetup{font=scriptsize}
    \centering
        \begin{minipage}{.45\linewidth}
            \begin{center}
            \captionof{table}{Systematic search for the best stopping step $s$ for the condition truncation.}
            \label{tab_fusion_cond_truncation}
            \scalebox{0.55}{
            \begin{tabular}{ccccccccc}
                \toprule
                Metric/Steps ($s$) & 0 & 10 & 20 & 40 & 80 & 120 & 160 & 200 \\
                \cmidrule(r){1-1}
                \cmidrule(r){2-9}
                $2^{nd}$ stage FID $\downarrow$ & \textbf{6.1} & 7.4 & 7.9 & 8.6 & 9.9 & 10.4 & 10.8 & 11.2 \\
                Overall FID $\downarrow$ & 11.6 & 10.7 & \textbf{9.5} & 10.9 & 11.2 & 13.5 & 15.7 & 18.9 \\
                \bottomrule
                \end{tabular}
            }
            \end{center}
        \end{minipage}
        \hfill
        \begin{minipage}{.5\linewidth}
            \begin{center}
                \captionof{table}{Ablation study of the sketch augmentation effect on the overall performance after fusing the abstract and the detailed stages.}
                \label{tab_fusion_aug}
                \scalebox{0.6}{
                \begin{tabular}{ccccc}
                    \toprule
                    \- & \multicolumn{1}{c}{\multirow{2}{*}{{\makecell{Cutout \cite{devries2017improved} \\ Percentage}}}}
                    & \multicolumn{1}{c}{\multirow{2}{*}{{\makecell{Dropout\\ Percentage}}}}
                    & \multicolumn{1}{c}{\multirow{2}{*}{{\makecell{$2^{nd}$ Stage\\ FID $\downarrow$}}}}
                    & \multicolumn{1}{c}{\multirow{2}{*}{{\makecell{Overall\\ FID $\downarrow$}}}}\\
                    \- & \- & \- & \- & \- \\
                    \cmidrule(r){1-3}
                    \cmidrule(r){4-5}
                    a) & 0    & 0         & 8.6  & 16.10 \\
                    b) & 5-10  & 5-20     & 9.89 & 13.94 \\
                    c) & 10-20 & 20-40    & 9.77 & 13.98 \\
                    d) & 20-30 & 50-70    & 9.80 & \textbf{13.76} \\
                    e) & 30-40 & 70-90    & 11.68& 17.99 \\
                    \bottomrule
                    \end{tabular}
                }
                \end{center}
        \end{minipage}
\end{figure}
%%%%%%%%%%%%%%%%%%%%%%%%    Figure      %%%%%%%%%%%%%%%%%%%%

The reported FID scores for the $1^{st}$ and the $2^{nd}$ stages, in Figure \ref{fig_qualitative_1st} and Figure \ref{fig_2nd_multimodal_ablation}, respectively, are for each stage separately.
In other words, in Figure \ref{fig_2nd_multimodal_ablation}, row c, the $2^{nd}$ stage achieves an 8.6 FID score when a GT sketch is fed, which is obtained from PidiNet \cite{su2021pixel}.
However, when we fed the generated sketch from the $1^{st}$ stage, the performance drastically dropped from 8.6 to 16.1 (almost doubled), as shown in Table \ref{tab_fusion_aug}, row a, due to the domain gap between the generated and the GT sketches.
To fill this gap, we explored two types of augmentations: 
1) Cutout and Dropout augmentation.
2) Condition truncation augmentation.

% --------------------------------------------------------
\noindent \textbf{Cutout and Dropout Augmentation.}
First, we explored the straightforward augmentation types, such as Cutout \cite{devries2017improved} and Dropout augmentations.
For the Cutout \cite{devries2017improved}, we apply a kernel to randomly blackout patches in the sketch.
Additionally, regarding the Dropout augmentation, we randomly convert white pixels to black pixels, interpreted as dropping some white points from the sketch.
As shown in Table \ref{tab_fusion_aug}, generally, augmentation leads to a drop in the $2^{nd}$ stage performance while helping fill the gap in the overall performance, as the FID score decreased from 16 to almost 14.
However, as shown in rows b-d, increasing the amount of the applied augmentation gradually does not help much, as the performance remains almost the same.
However, the overall accuracy, i.e., FID, drops significantly from 14 to 18 when an aggressive augmentation is applied.

% --------------------------------------------------------
\noindent \textbf{Condition Truncation Augmentation.}
As shown in Table \ref{tab_fusion_aug}, the conventional augmentation techniques do not help much. 
Accordingly, we explored another augmentation variant tailored for the diffusion models, i.e., condition truncation \cite{ho2022cascaded}.
During training the $2^{nd}$ stage, we apply a random Gaussian noise to the fed condition, i.e., the sketch. 
So now the $2^{nd}$ stage is trained on noisy sketches instead of pure ones, which makes it more robust to the variations between the real sketches and the generated ones from the $1^{st}$ stage.
Typically, we progressively generate the sketch in $T$ time steps; $T \rightarrow 0$. 
However, This added noise could be interpreted as we stop the sketch generation ($1^{st}$ stage) at a particular step $s$; $T \rightarrow s$.
Consequently, we search for it during sampling to determine which step $s$ works best for the overall performance, as shown in Table \ref{tab_fusion_cond_truncation}.
In other words, the $2^{nd}$ stage is trained on a wide range of perturbed sketches, e.g., pure sketch ($s=0$), noisy one ($0<s<T$), and even pure noise ($s=T$).
The $1^{st}$ stage is trained for 200 steps, so $s=200$ means we omit the sketch and feed pure noise. In contrast, $s=0$ indicates that we feed the generated sketch as it is to the $2^nd$ stage.
As shown in Table \ref{tab_fusion_cond_truncation}, following this truncation mechanism, the fusion of the two stages is performed more efficiently, where the overall performance drops from 16.1 to 10.6.
Algorithm \ref{alg_sampling_pipeline} is adapted in the supplementary materials to incorporate the condition truncation mechanism.

% -----------------------------------------------------------
% \clearpage
\section{More qualitative results}

% ----------------------------------------------------------
\subsection{Demo}
\label{sec_appendix_demo}

Please refer to our \href{https://toddlerdiffusion.github.io/website/}{$demo$}, which probes our controllability capabilities in performing consistent edits.

Our demo demonstrates a lot of use cases, including:
\begin{itemize}
    \item (0:00 --> 0:20) [1st stage] Starting from noise generates a sketch by running only the 1st stage, which is unconditional generation.
    \item (0:20 --> 0:24) [2nd stage] Starting from a generated sketch generates a palette. This one is considered an unconditional generation as noise generates the sketch unconditionally.
    \item (0:24 --> 0:32) [3rd stage] Starting from a generated palette generates an RGB image. This one is also considered an unconditional generation.
    \item (0:32 --> 0:44) [Conditional Generation] Starting from the GT sketch, we generate an RGB image.
    \item (0:44 --> 1:15) [Editing] We show our sketch-based editing capabilities.
    \item (1:15 --> 1:46) [Editing] We show our palette-based editing capabilities.
\end{itemize}

% ----------------------------------------------------------
% ----------------------------------------------------------
\subsection{Real Image Editing}
\label{sec_appendix_rela_editing_SEEDx}

As explained in Section \ref{sec_toddler_consistency}, we ensure consistent edits for real images by storing the noise used during the generation process for later edits. The primary challenge with real image editing is that we don’t have access to the noise values at each denoising step, as our model didn’t generate the image. To address this, we employ DDPM-inversion \cite{huberman2024edit} to reconstruct the intermediate noise. 
As illustrated in Figure \ref{fig_real_img_editing_vs_seed}, we compare our model against SEED-X \cite{ge2024seed}, and our model achieves consistent edits on real reference images better than SEED-X, despite being trained on much fewer data compared to SEED-X and using smaller architecture.
Additionally, SEED-X fails to perform the edits correctly and keeps the rest of the image unedited, such as the rest of the face or the background.
On the contrary, our method shows great faithfulness to the original image while performing the edits depicted in the sketch.

\input{Figures/appendix/real_img_editing_vs_SEED}

% ----------------------------------------------------------
\subsection{Class-Label Robustness}
\label{sec_appendix_imgnet_robust}

Beyond these quality comparisons, we also conducted a robustness analysis to assess our model's behavior when faced with inconsistencies between conditions. 
Specifically, we deliberately provided incorrect class labels for the same sketch.
This mismatched the sketch (geometry) and the desired class label (style). Despite the inconsistent conditions, as illustrated by the off-diagonal images in Figure \ref{fig_imgnetRobust_full}, our model consistently adheres to the geometry defined by the sketch while attempting to match the style of the misassigned category, demonstrating the robustness of our approach. Successful cases are highlighted in yellow, and even in failure cases (highlighted in red), our model avoids catastrophic errors, never hallucinating by completely ignoring either the geometry or the style.

\input{Figures/appendix/imgnet_robustness_full}

% ----------------------------------------------------------
\subsection{Editing Capabilities}
\label{sec_More_qualitative_results_for_editing_capabilities}

\begin{figure}[H]
\centering
\includegraphics[width=0.99\linewidth]{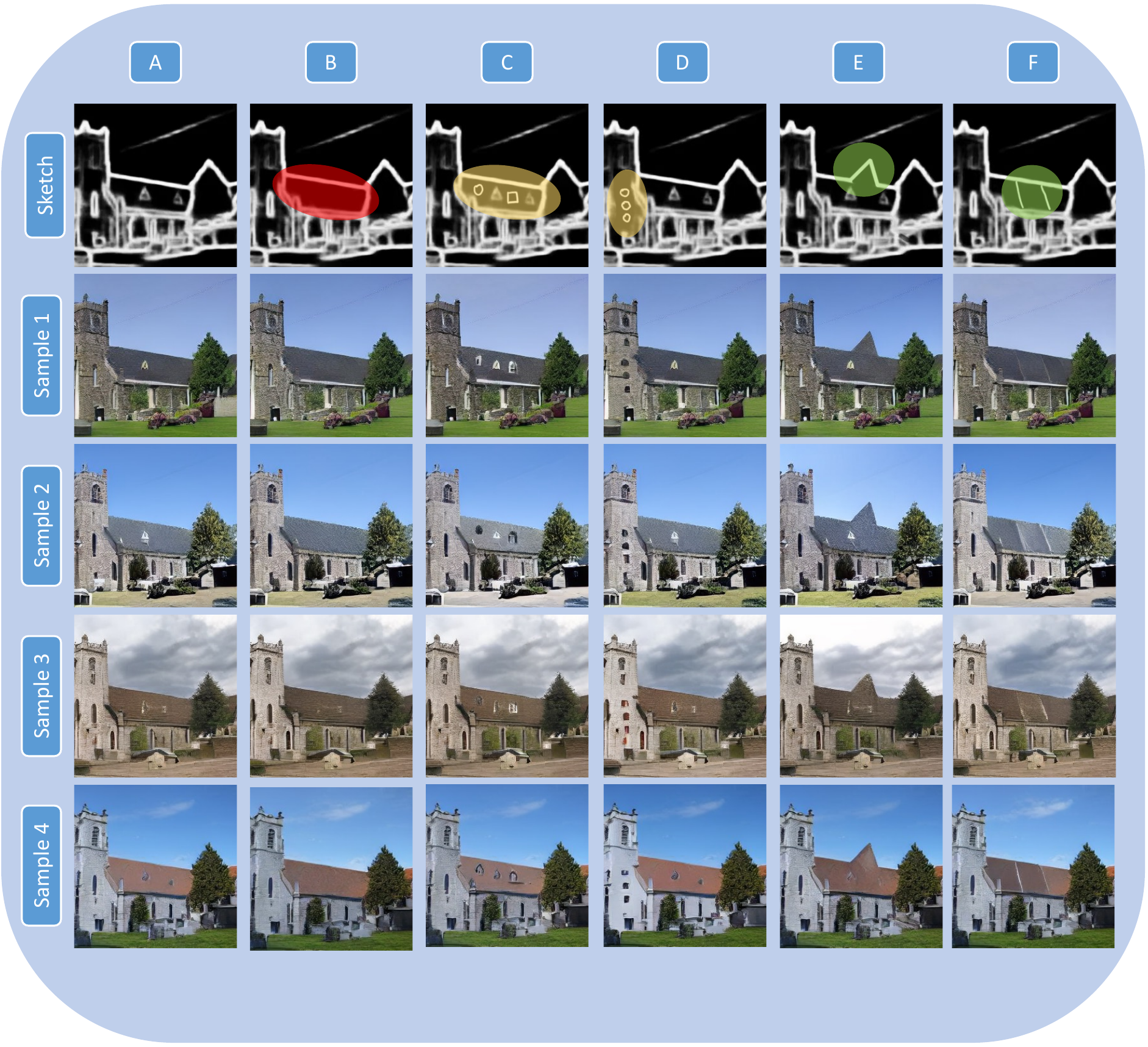}
\caption{Controllability ability of our framework, {\papernameAbbrev}.
Starting from generated sketch and RGB image (A), we can remove artifacts or undesired parts, in red, (B), add a new content, in yellow, (C-D), and edit the existing content, in green, (E-F) by manipulating the sketch.}
\label{fig_controlability_qualitative_1}
\end{figure}

\begin{figure}[H]
\centering
\includegraphics[width=0.99\linewidth]{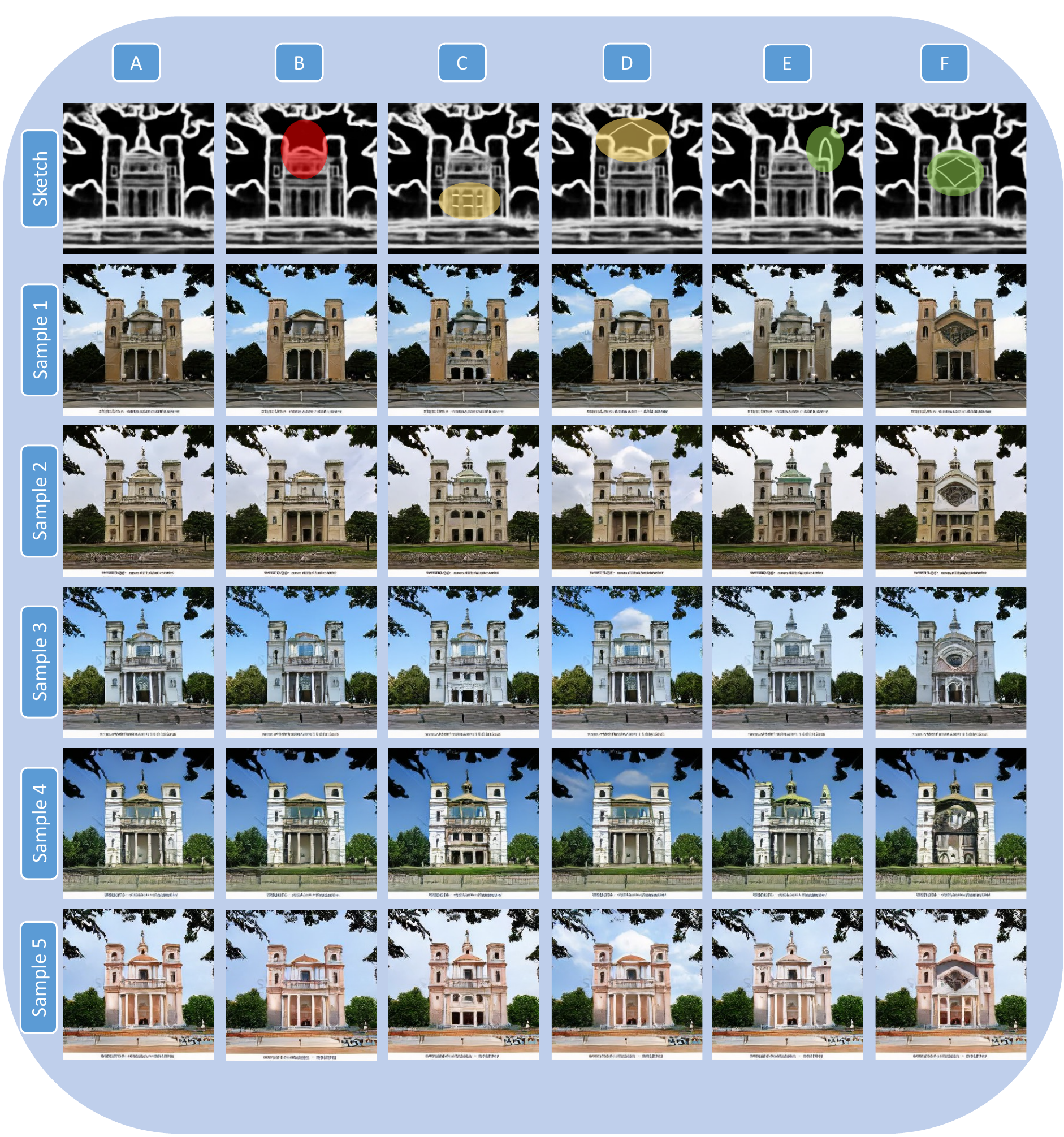}
\caption{Controllability ability of our framework, {\papernameAbbrev}.
Starting from generated sketch and RGB image (A), we can remove artifacts or undesired parts, in red, (B), add a new content, in yellow, (C-D), and edit the existing content, in green, (E-F) by manipulating the sketch.}
\label{fig_controlability_qualitative_2}
\end{figure}

\begin{figure}[H]
\centering
\includegraphics[width=0.99\linewidth]{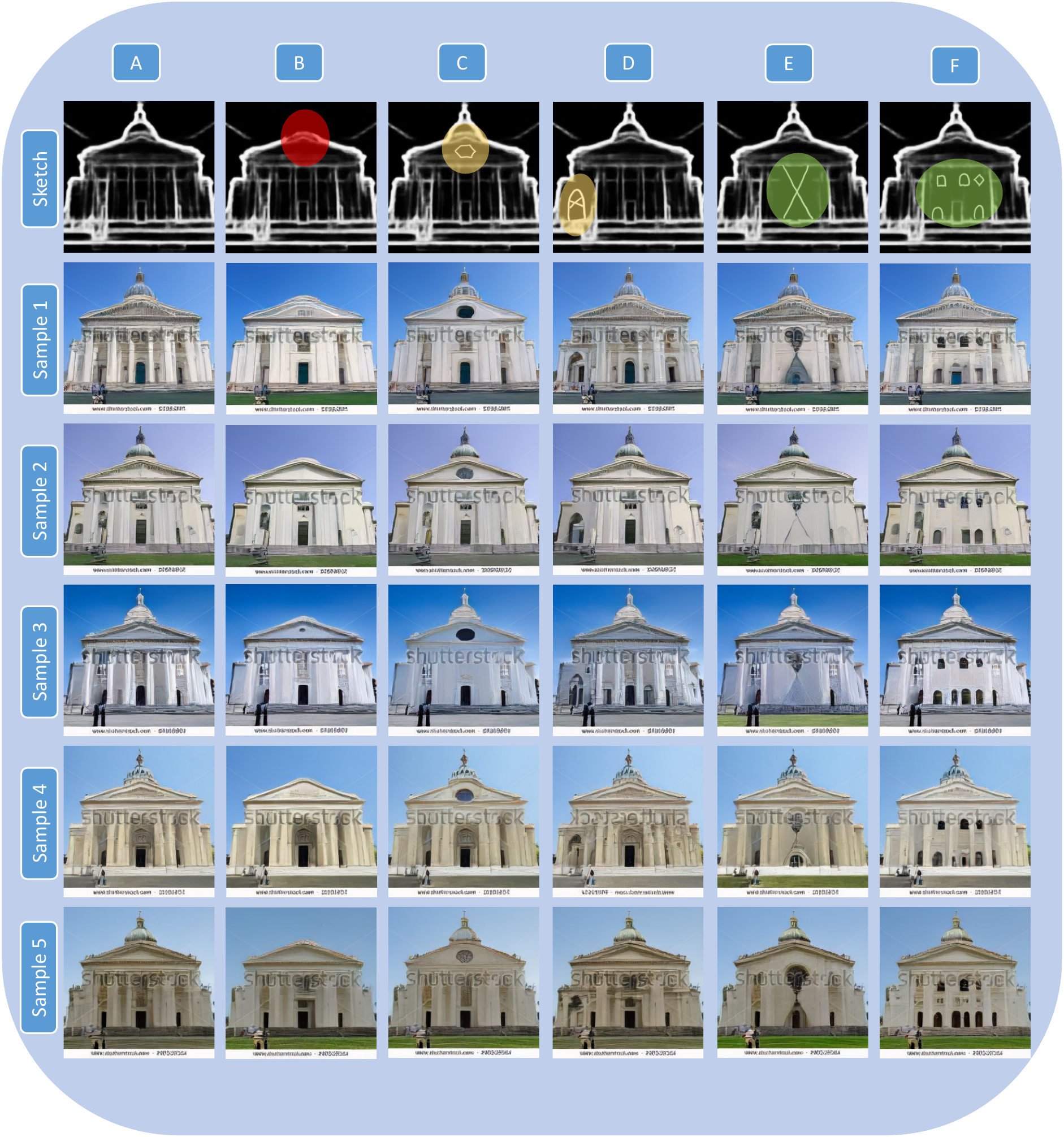}
\caption{Controllability ability of our framework, {\papernameAbbrev}.
Starting from generated sketch and RGB image (A), we can remove artifacts or undesired parts, in red, (B), add a new content, in yellow, (C-D), and edit the existing content, in green, (E-F) by manipulating the sketch.}
\label{fig_controlability_qualitative_3}
\end{figure}

\begin{figure}[H]
\centering
\includegraphics[width=0.99\linewidth]{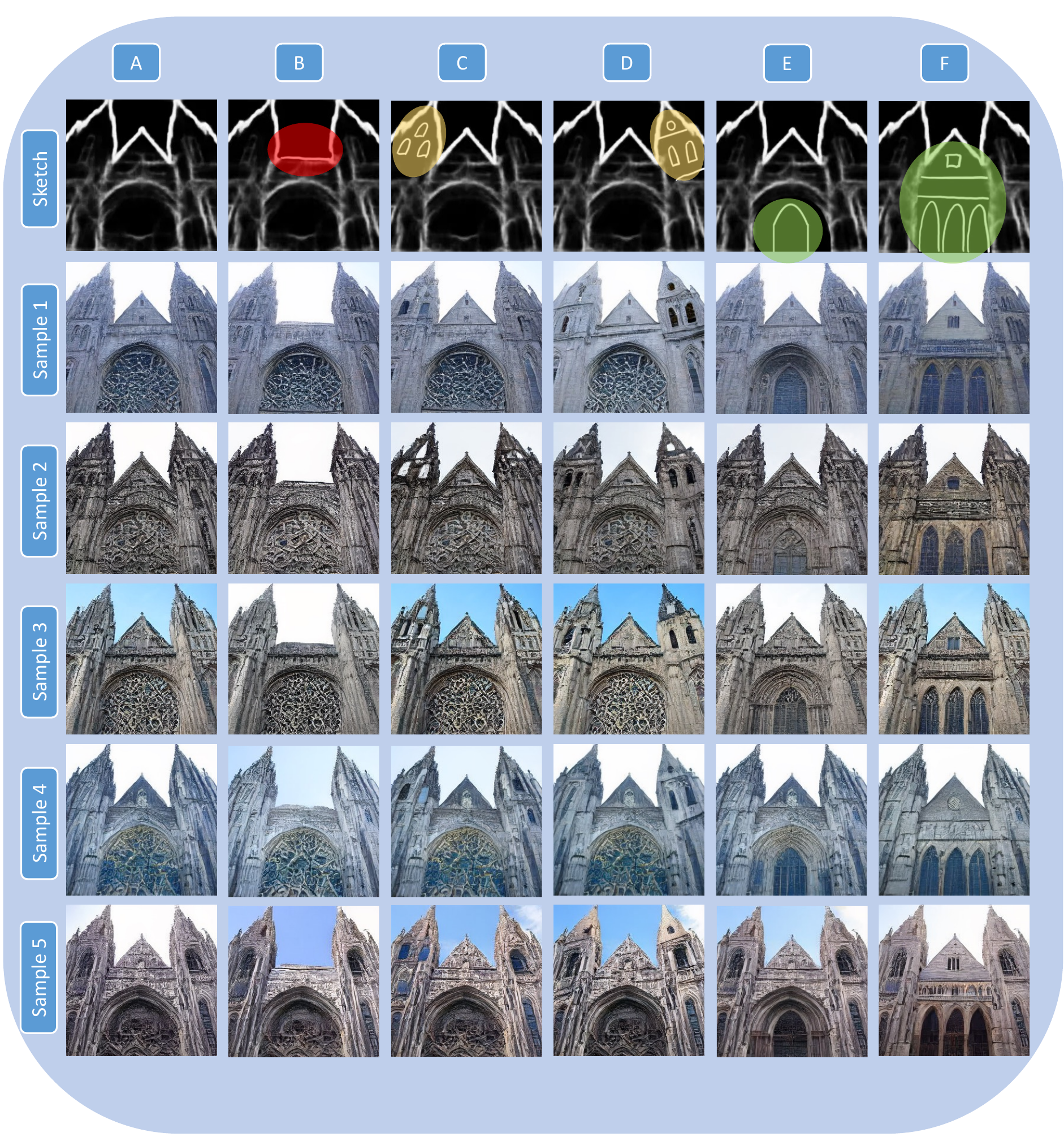}
\caption{Controllability ability of our framework, {\papernameAbbrev}.
Starting from generated sketch and RGB image (A), we can remove artifacts or undesired parts, in red, (B), add a new content, in yellow, (C-D), and edit the existing content, in green, (E-F) by manipulating the sketch.}
\label{fig_controlability_qualitative_4}
\end{figure}

Our approach introduces an interpretable generation pipeline by decomposing the complex RGB generation task into a series of interpretable stages, inspired by the human generation system \cite{stevens2012vision, marr2010vision, marr1974purpose}. 
Unlike traditional models that generate the complete image in one complex stage, we break it into $N$ simpler stages, starting with abstract contours, then an abstract palette, and concluding with the detailed RGB image. This decomposition not only enhances interpretability but also facilitates dynamic user interaction, offering unprecedented editing capabilities for unconditional generation, as shown in Figure \ref{fig_controlability_qualitative_1}, Figure \ref{fig_controlability_qualitative_2}, Figure \ref{fig_controlability_qualitative_3}, and Figure \ref{fig_controlability_qualitative_4}. 

% -----------------------------------------------------------
% -----------------------------------------------------------
% -----------------------------------------------------------
% -----------------------------------------------------------
% -----------------------------------------------------------
% -----------------------------------------------------------
% -----------------------------------------------------------
% -----------------------------------------------------------

% ---------------------------------------------------------------------------
\subsection{Sketch Conditioning (SD-v1.5)}
\label{sec_appendix_Sketch Conditioning(SD-v1.5)}

We convert our method to a ``sketch2RGB'' by omitting the $1^{st}$ stage to show our controllability capabilities.
By doing this, we can compare our model against the vast tailored sketch-based models built on top of SD-v1.5.
However, these methods have been trained for a long time on large-scale datasets.
Thus, for a fair comparison, we retrain them using CelebHQ datasets for 5 epochs starting from SD-v1.5 weights.
As demonstrated in Figure \ref{fig_Controlability_SD_benchmark}, despite this is not our main focus, our method outperforms conditioning methods without adding additional modules \cite{zhang2023adding, zhao2024uni} or adapters \cite{mou2024t2i}.
We compare training-free \cite{tumanyan2023plug, parmar2023zero} and training-based \cite{li2023gligen, zhang2023adding, zhao2024uni} methods.
The performance is worse than expected for the trained-based methods due to two reasons:
1) We train all the models for only 5 epochs; thus, some of this method, in contrast to us, needs more epochs to converge.
2) We apply augmentations to the input sketch, which shows that some of these methods are vulnerable to sketch perturbations. 

\input{Figures/appendix/control_SD_benchmark}

\clearpage
% ---------------------------------------------------------------------------
\subsection{Free-Hand Sketch Conditional Generation Samples}
\label{sec_Qualitative_results_for_our_unconditional_generation}

\begin{figure}[H]
\centering
\includegraphics[width=0.99\linewidth]{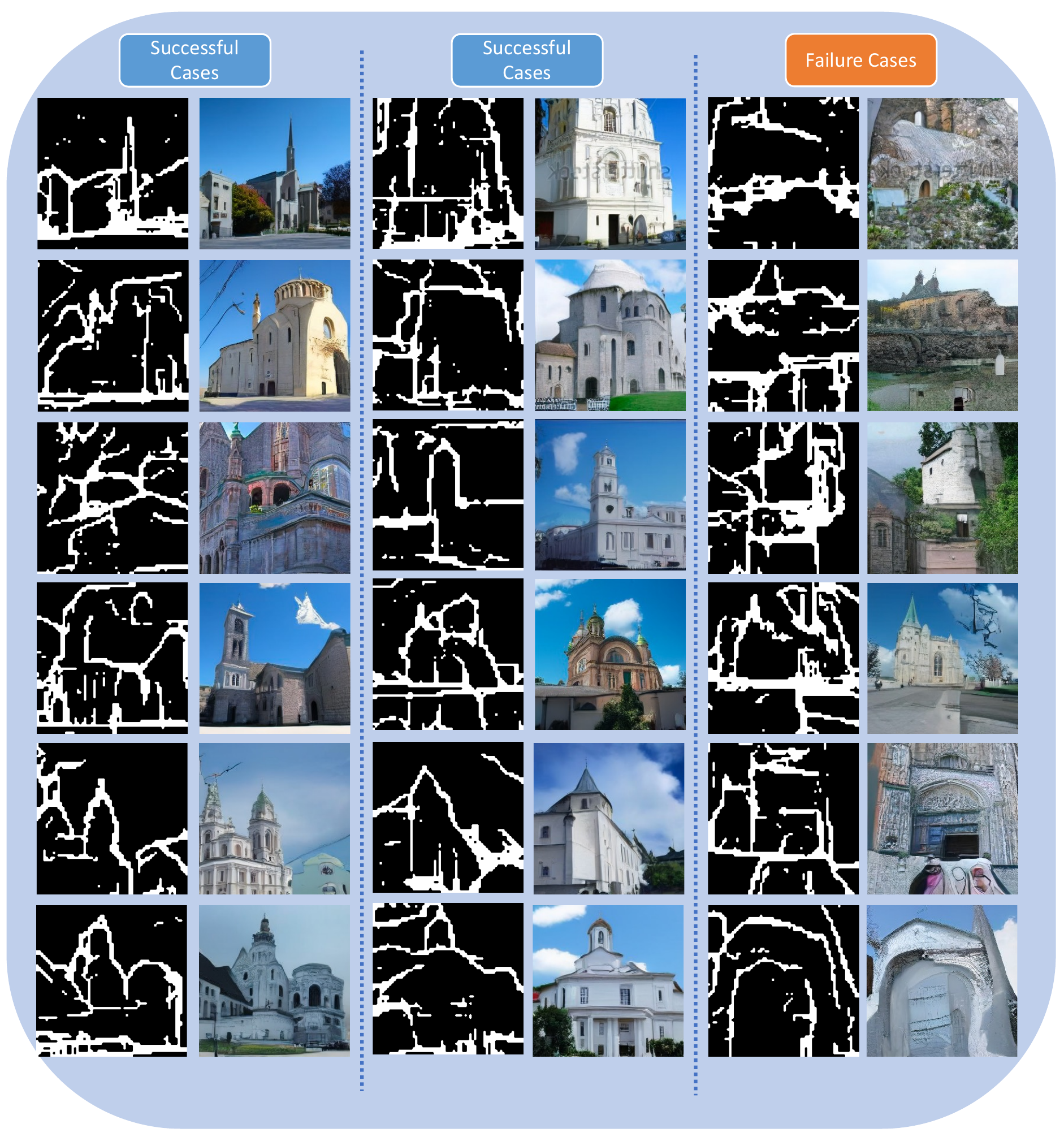}
\caption{Qualitative results for our unconditional generation.}
\label{fig_uncoditional_qualitative}
\end{figure}

Figure \ref{fig_uncoditional_qualitative} depicts more qualitative results for our conditional generation pipeline, where we first generate the sketch using drawing tools and then use the generated sketch to generate the RGB image.
Despite the sketch being too bad and sparse, our model still generates high-quality images.
% -----------------------------------------------------------
% -----------------------------------------------------------

\clearpage
% ---------------------------------------------------------------------------
\subsection{Qualitative results for our sketch-conditional generation}
\label{sec_Qualitative results for our sketch-conditional generation}

\begin{figure}[H]
\centering
\includegraphics[width=0.99\linewidth]{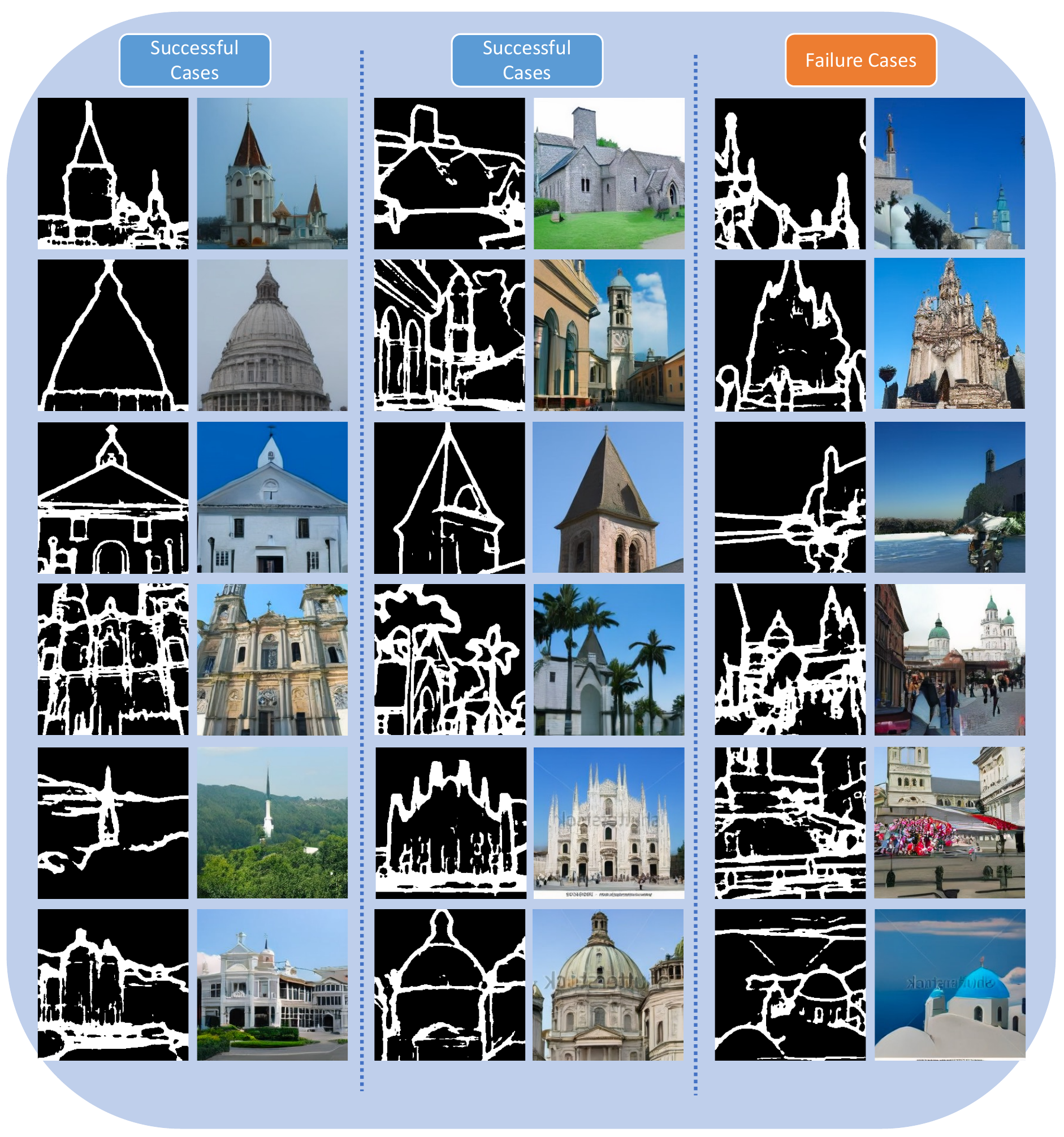}
\caption{Qualitative results for our sketch-conditional generation.}
\label{fig_sketch_coditional_qualitative}
\end{figure}

In addition to our unconditional generation, our framework can be used as a conditional generation pipeline, where the user can omit the first stage and directly fed a reference sketch.
As shown in Figure \ref{fig_sketch_coditional_qualitative}, given the reference sketch, we run only the second stage, which is responsible for generating the RGB image given a sketch.
Compared to the generated images in Figure \ref{fig_uncoditional_qualitative}, the generated images in Figure \ref{fig_sketch_coditional_qualitative} are much better as the reference sketches are much better than the generated ones.
This observation is aliened with our discussion regarding the gap between the two stages.
Therefore, improving the quality of the generated sketches will directly reflect on the overall images quality.
% -----------------------------------------------------------
% -----------------------------------------------------------

\clearpage
% ---------------------------------------------------------------------------
\section{Failure cases}
\label{sec_Failure cases}

The last column of Figure \ref{fig_uncoditional_qualitative} and Figure \ref{fig_sketch_coditional_qualitative} depicts failure cases, where the generated RGB is following the sketch but generates unrealistic image.
Intuitively, the failure cases are more apparent in the unconditional generation due to the generated sketches quality.

%% file: Figures/tex/arch_horizontal.tex
% --------------------------------------------------------
\begin{figure}[ht!]
% \captionsetup{font=scriptsize}
\centering
\includegraphics[width=0.99\linewidth]{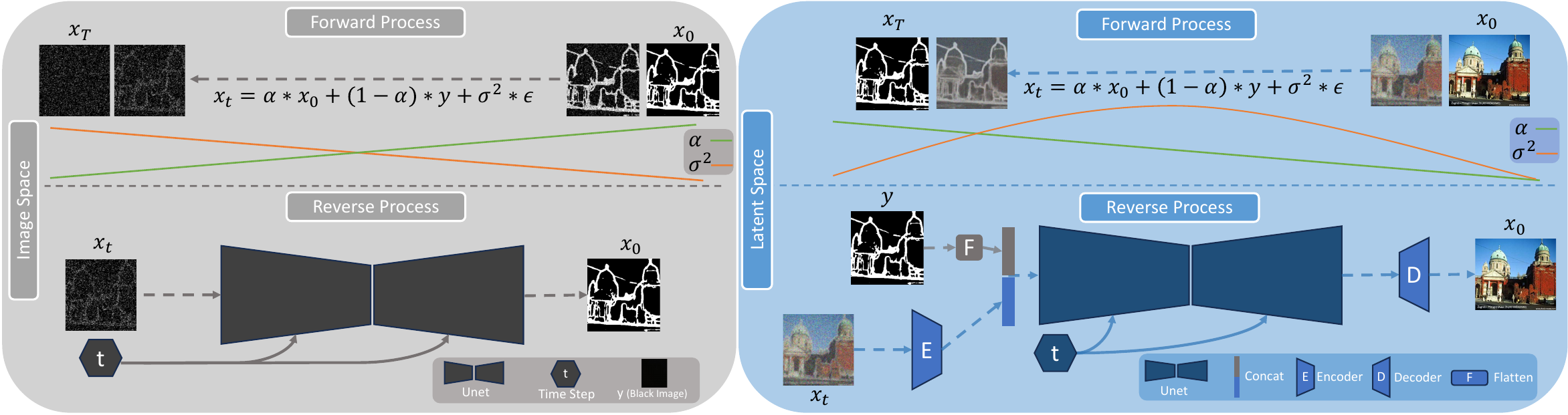}
\caption{An overview of proposed architecture, dubbed {\papernameAbbrev}. The first block demonstrates the first stage, which generates a sketch unconditionally. Due to our efficient formulation, this stage operates in the image space on $64 \times 64$ resolution. The bottom module depicts the third stage, which generates an RGB image given a sketch only or both sketch and palette.}
\label{fig_cot_arch}
% \vspace{-1.5em}
\end{figure}
% --------------------------------------------------------

%% file: Figures/7_1ststage_results.tex
% --------------------------------------------------------
\begin{figure}[t!]
\captionsetup{font=scriptsize}
\begin{minipage}[c]{.5\linewidth}
    %\vspace{0pt}
    \begin{center}
        \includegraphics[width=0.9\linewidth]{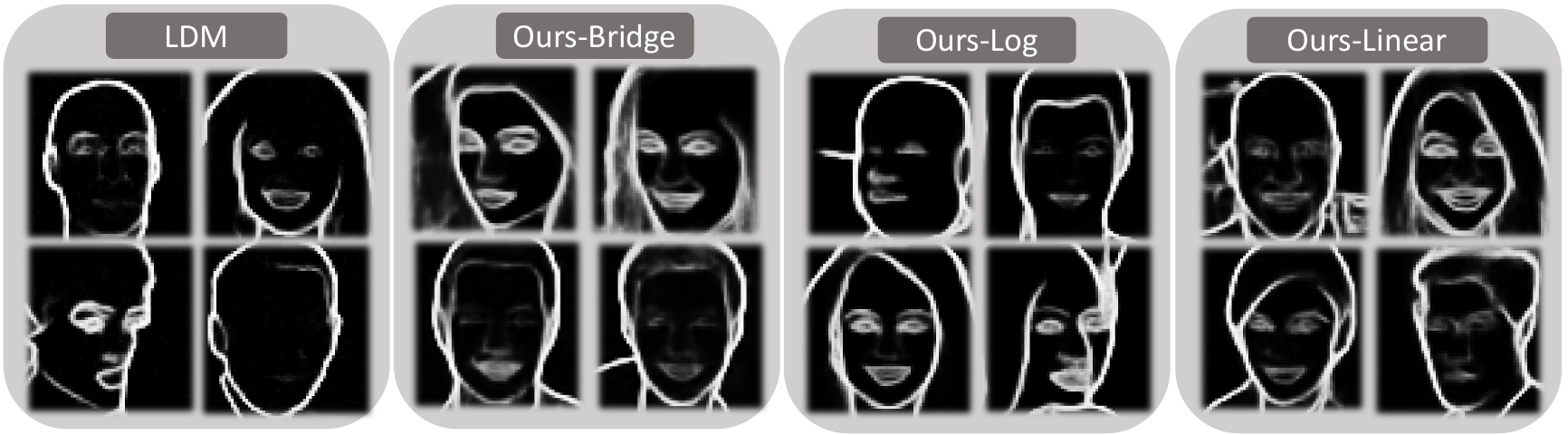}
    \end{center}
\end{minipage}
\begin{minipage}[c]{.5\linewidth}
    \begin{center}
    \scalebox{0.5}{
    \begin{tabular}{cccccccc}
        \toprule
        \- & \multicolumn{1}{c}{\multirow{3}{*}{{Method}}}
        & \multicolumn{1}{c}{\multirow{3}{*}{{\makecell{Unet\\ Parameters}}}}
        & \multicolumn{1}{c}{\multirow{2}{*}{{\makecell{Training\\ Time}}}}
        & \multicolumn{1}{c}{\multirow{2}{*}{{\makecell{Sampling\\ Time}}}}
        & \multicolumn{1}{c}{\multirow{3}{*}{{\makecell{Noise\\ Scheduler}}}}
        & \multicolumn{1}{c}{\multirow{3}{*}{{\makecell{Sketch\\ FID $\downarrow$}}}}
        & \multicolumn{1}{c}{\multirow{3}{*}{{\makecell{RGB\\ FID $\downarrow$}}}} \\
        \- & \- & \- & \- & \- & \- & \- & \- \\
        \- & \- & \- & (Min:Sec)$\downarrow$ & (FPS)$\uparrow$ & \- & \- & \- \\
        \cmidrule(r){1-6}
        \cmidrule(r){7-8}
        a) & LDM  & 187 M & 4:05 & 2.2 & Linear   & 126.4 & 23.5  \\
        b) & Ours     &  1.9 M & 0:18 & 53 & Bridge                   & 110.5 & 35.2 \\
        c) & Ours     &  1.9 M & 0:18 & 53 & Log                      &  61.2 & 15.6 \\
        d) & Ours     &  \textbf{1.9 M} & \textbf{0:18} & \textbf{53} & Linear          &  \textbf{58.8} & \textbf{15.1} \\
        \bottomrule
        \end{tabular}
    }
    \end{center}
\end{minipage}
\caption{Benchmarking results for the $1^{st} stage$ on the CelebA-HQ dataset \cite{karras2017progressive}. The training time is calculated per epoch using 4 NVIDIA RTX A6000. The sampling time is calculated per frame using one NVIDIA RTX A6000 with a batch size equals 32.}
\label{fig_qualitative_1st}
\vspace{-3mm}
\end{figure}
% --------------------------------------------------------

%% file: Figures/appendix/paletteFig_and_sketchTable.tex
% --------------------------------------------------------
\begin{table}[t!]
\captionsetup{font=scriptsize}
\begin{minipage}[c]{.55\linewidth}
    %\vspace{0pt}
    \centering
    \captionof{table}{Sketch Intensity Analysis.}
    \scalebox{0.7}{
    \begin{tabular}{cccccc}
        \toprule
        \textbf{Dataset/Intensity} &  \textbf{19\%} &  \textbf{24\%} & \textbf{31\%} & \textbf{37\%} & \textbf{43\%}  \\ 
        \midrule
        CelebHQ & \textbf{7.4} & 7.9 & 9.2 & 10.5 & 12.7 \\
        Lsun-Churches & 7.8 & \textbf{6.9} & 8.3 & 8.9 & 10.1 \\
        \bottomrule
    \end{tabular}
    \label{tab_intensity_ablation}
    }
\end{minipage}
\hfill
\begin{minipage}[c]{.4\linewidth}
    \centering
    \includegraphics[width=0.8\linewidth]{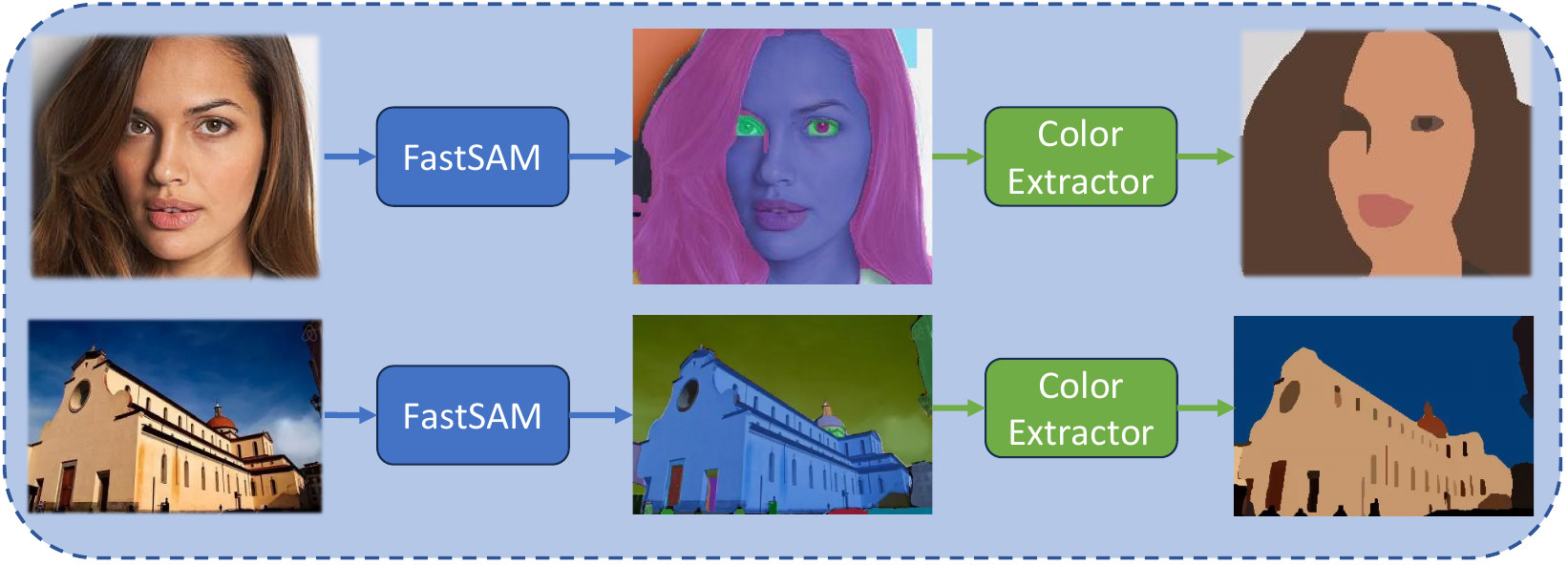}
    \captionof{figure}{Palette generation pipeline. First, we employ FastSAM \cite{zhao2023fast} to segment the image. Then,  we get the dominant color for each segment.}
    \label{fig_palette_pipeline}
\end{minipage}
\end{table}
% --------------------------------------------------------

%% file: Figures/appendix/real_img_editing_vs_SEED.tex
\begin{figure}[H]
\centering
\includegraphics[width=0.99\linewidth]{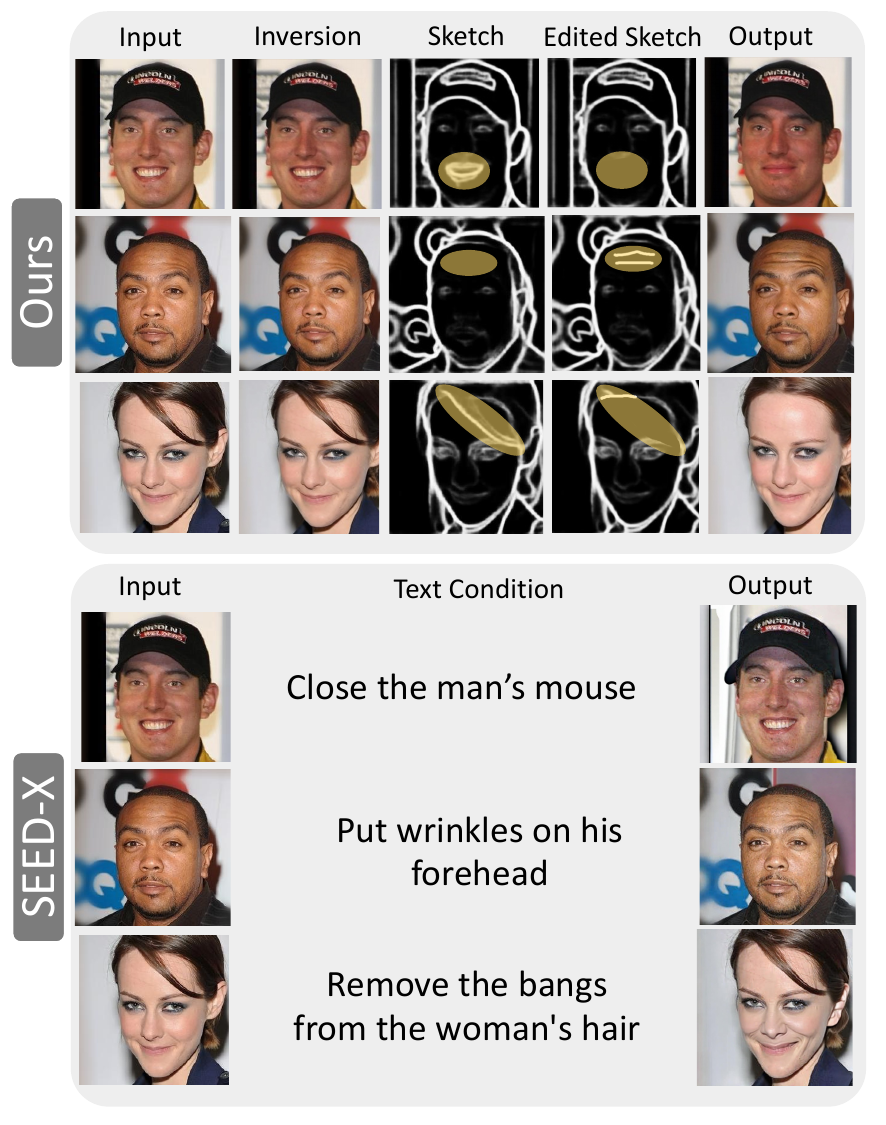}
\caption{Comparison between our approach, {\papernameAbbrev} and SEED-X \cite{ge2024seed} in terms of the real images capabilities.}
\label{fig_real_img_editing_vs_seed}
\end{figure}
% -----------------------------------------------------------

%% file: Figures/appendix/imgnet_robustness_full.tex
\begin{figure}[H]
\centering
\includegraphics[width=0.99\linewidth]{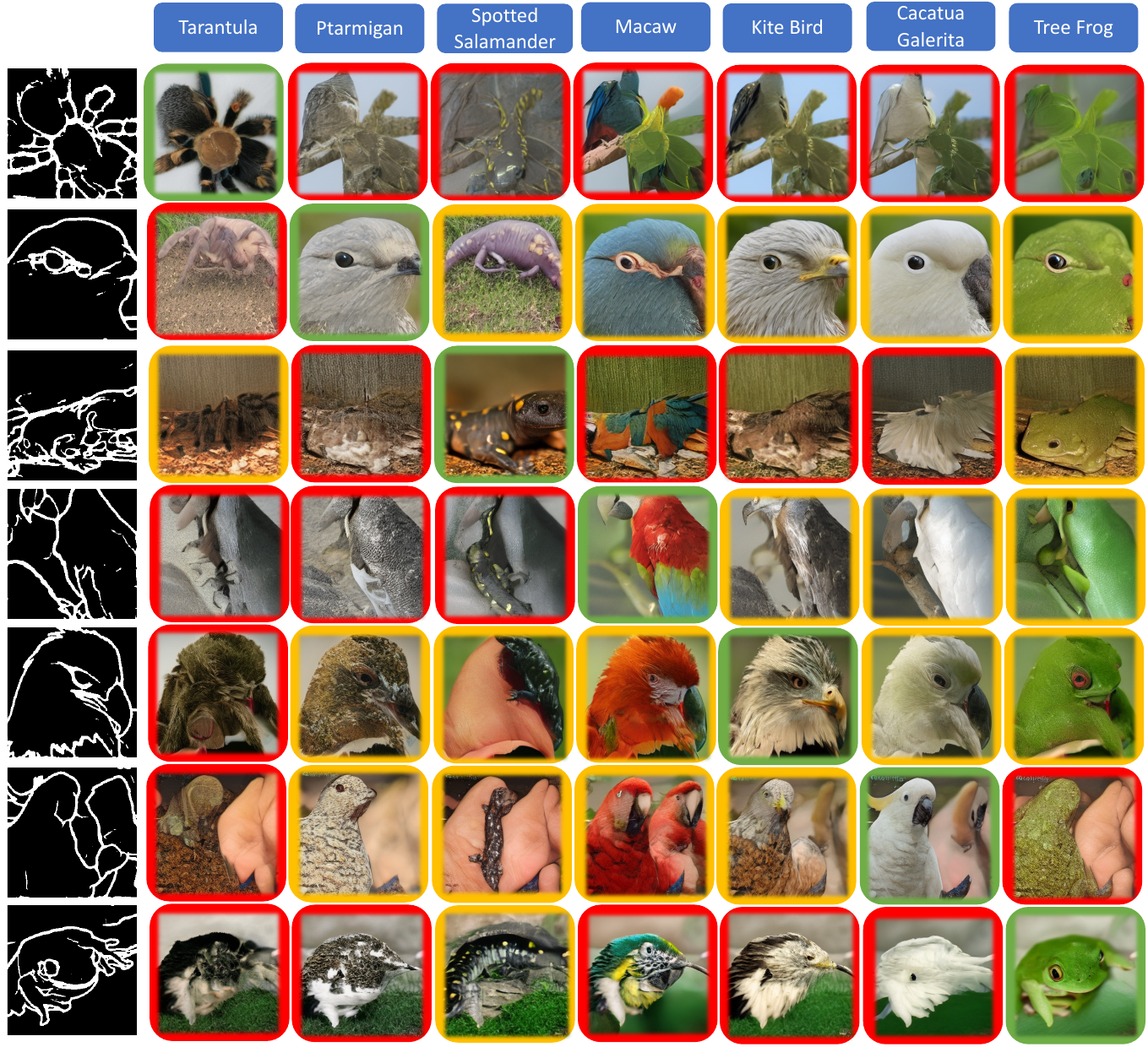}
\caption{Our results on the ImageNet dataset. The first column depicts the input sketch. The diagonal (green) represents the generated output that follows the sketch and the GT label. The off-diagonal images show the output given the sketch and inconsistent label. These adversarial labels show our model's robustness, where the successful cases are in yellow, and the failure ones are in red.}
\label{fig_imgnetRobust_full}
\end{figure}
% -----------------------------------------------------------

%% file: Figures/appendix/control_SD_benchmark.tex
\begin{figure}[H]
\centering
\includegraphics[width=0.99\linewidth]{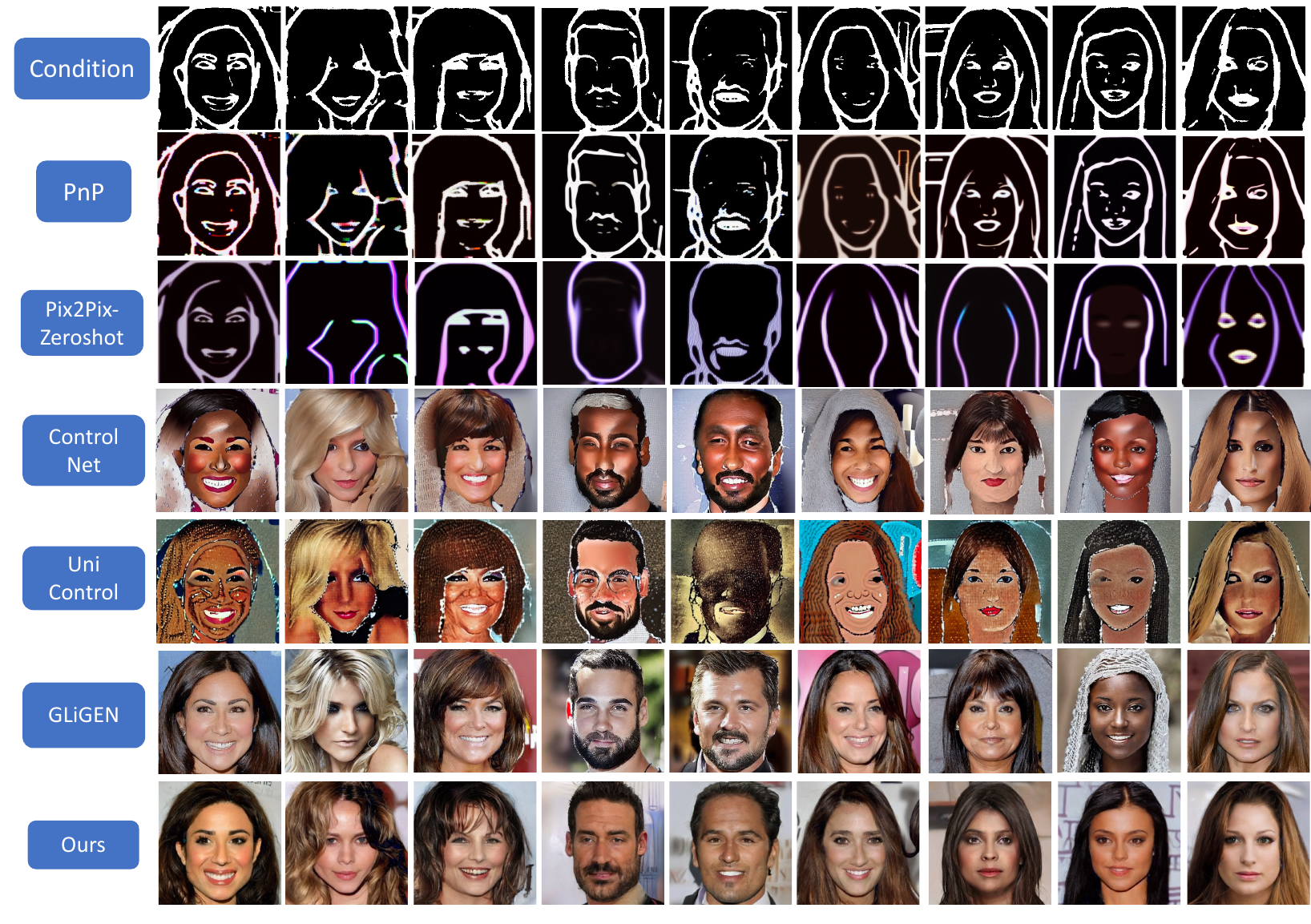}
\caption{Comparison between our approach, {\papernameAbbrev} and training-free \cite{tumanyan2023plug, parmar2023zero} and training-based \cite{li2023gligen, zhang2023adding, zhao2024uni} sketch conditioning methods.}
\label{fig_Controlability_SD_benchmark}
\end{figure}
% -----------------------------------------------------------